\documentclass{article}

\usepackage[preprint]{neurips_2026}

\usepackage{float}
\usepackage{graphicx} 
\usepackage{amsfonts}
\usepackage{amssymb}
\usepackage{latexsym}
\usepackage{mathrsfs}
\usepackage{amsmath}
\usepackage{amsthm}
\usepackage{xcolor}
\usepackage{algorithm}
\usepackage{algpseudocode}

\usepackage{amssymb}
\usepackage{enumerate}
\usepackage{CJKutf8}
\usepackage{caption}
\usepackage{subcaption}
\usepackage{overpic}  
\usepackage{booktabs}
\usepackage{url}
\usepackage[utf8]{inputenc}
\usepackage{tikz}
\usepackage{wrapfig}
\usepackage{svg}
\usepackage{multirow}
\usepackage{caption}
\usepackage{booktabs}
\usepackage{changepage}
\usetikzlibrary{arrows.meta, positioning, shapes.geometric}
\usepackage[colorlinks=true, linkcolor=blue, citecolor=blue, urlcolor=blue]{hyperref}

\newtheorem{theorem}{Theorem}[section]
\newtheorem{lemma}[theorem]{Lemma}
\newtheorem{definition}[theorem]{Definition}

\newtheorem{corollary}[theorem]{Corollary}

\newtheorem{proposition}[theorem]{Proposition}

\newcommand*{\diam}{\mathrm{diam}}
\newcommand*{\dimb}{\dim_{\mathrm B}}
\newcommand*{\tdimb}{\widetilde{\dim}_{\mathrm B}}

\title{Fractal Graph Contrastive Learning}
\author{%
  Nero Z. Li$^*$ \\
  Department of Mathematics \\
  Imperial College London \\
  CDT, University of Oxford \\
  \texttt{ziyu.li21@imperial.ac.uk} \\
  \And
  Xuehao Zhai\thanks{equal contribution} \\
  Department of Civil and Environmental Engineering\\
  Imperial College London \\
  \texttt{x.zhai20@imperial.ac.uk} \\
  \And
  Zhichao Shi \\
  IDEA Research, International Digital Economy Academy \\
  School of Advanced Interdisciplinary Sciences, UCAS
  \\
  State Key Lab of AI Safety,
  Institute of Computing Technology, CAS \\
  \texttt{shizhichao22@mails.ucas.ac.cn} \\
  \And
  Boshen Shi \\
  China Mobile Research Institute \\
  \texttt{boshenshi97@gmail.com}
  \And
  Xuhui Jiang\thanks{corresponding author}\\
  DataArc Tech Ltd. \\
  IDEA Research,
  International Digital Economy Academy \\
  \texttt{jiangxuhui@idea.edu.cn} \\
}
\date{}

\begin{document}

\maketitle

\begin{abstract}
Graph Contrastive Learning (GCL) relies on semantically consistent graph augmentations, but common local perturbations provide limited control over global structural consistency, motivating a more principled global augmentation strategy.
We therefore propose Fractal Graph Contrastive Learning (FractalGCL), a theory-motivated framework that constructs a renormalisation-based augmented graph and introduces a fractal-dimension-aware contrastive loss that penalises unreliable positive views and reweights negative-pair repulsion by finite-scale box-counting discrepancies.
However, computing these discrepancies introduces substantial overhead, so we derive and justify a Gaussian surrogate that avoids repeated box-counting on renormalised graphs, yielding about a $61\%$ runtime reduction.
Experiments show that FractalGCL serves as an effective frozen-pretraining tool on MalNet-Tiny, achieves strong performance on the standard TUDataset benchmarks, and outperforms the next-best method on real-world urban traffic tasks by $4.51$ percentage points in average accuracy.
Code is available at \url{https://anonymous.4open.science/r/FractalGCL-0511/}.

\end{abstract}


\section{Introduction}


Graph contrastive learning (GCL) has emerged as a popular self-supervised paradigm for graph representation learning~\citep{hu2020strategies,xia2022simgrace,you2021graph,ju2024towards,liu2022graph,xu2018powerful}.
By forcing models to discriminate positive pairs from negative pairs, it alleviates the knowledge scarcity problems~\citep{wu2021self,xie2022self,chen2025data,shi2025domain}, and it also serves as an effective pretext task for pre-training graph foundation models~\citep{liu2023towards,huang2024large}. As graphs possess non-Euclidean topology, researchers must tailor
contrastive learning frameworks to graph-specific properties. Therefore, GCL has formed unique lines of research, which cover stages including augmenting graph data~\citep{liu2022towards,rong2019dropedge,sun2021mocl}, designing contrastive modes~\citep{ju2023unsupervised,ren2021label,park2020unsupervised}, and optimizing contrastive objectives~\citep{hjelm2018learning,xia2022progcl,zhang2022localized}.

Among current studies, data augmentation remains a pivotal challenge in graph contrastive learning, as the quality of positive and negative sample pairs fundamentally determines the capacity of a graph model to extract meaningful knowledge and the quality of learned representations. While negative samples are typically generated by contrasting views from structurally distinct graphs or subgraphs~\citep{you2020graph}, which ensures divergent distributional characteristics, the generation of semantically coherent positive samples remains a critical bottleneck. Specifically, existing approaches often rely on random perturbations (e.g., node/edge deletion, attribute masking) or fixed topological constraints (e.g., hierarchy preservation), which provide only incomplete guarantees for maintaining structural consistency. These methods lack an explicit mechanism to ensure global similarity between the original graph and its augmented views, leading to potential mismatches in semantic alignment. This gap naturally raises a fundamental question: \emph{Can we design a principled graph-level criterion to enforce global structural consistency during positive sample generation?}

This critical question directs attention to a fundamental yet often overlooked global property of graphs—their inherent self‑similarity and hierarchical complexity, which is mathematically formalised through the concept of \textbf{fractal}. 
Fractal geometry~\citep{edgar2008measure,mandelbrot1983fractal,mandelbrot1989fractal} is a field of mathematics that explores irregular shapes whose intricate detail persists across different scales, appearing in patterns such as snowflakes, coastlines, and branching trees.
Fractal graphs are networks that possess fractal properties, effectively transplanting fractal concepts from Euclidean space onto graph structures (see Figure~\ref{fig:example}). 
Given the prevalence of fractal graphs in nature and society, their fractal properties likely play a significant yet under-explored role in improving graph representations via GCL.

\definecolor{myaqua}{RGB}{15,205,255}
\newcommand{\mynode}[2]{\begin{scope}[shift={(#1)},scale=#2]\node[node] (0,0) () {};\end{scope}}
\newcommand{\Rrule}[4]{\begin{scope}[shift={(#1)},scale=#2,rotate=#3]
  \foreach \nn/\x/\y in {0/0/0, 1/1/0, 2/1/1, 3/2/0, 4/2/1, 5/3/0}{\node[node] (\nn) at (\x,\y) {};}
  \draw[black] (0) -- (5) (1) -- (2) (3) -- (4);
  \foreach \nn in {0,...,5}{\node[node] () at (\nn) {};}#4\end{scope}}
\newcommand{\RRrule}[3]{\begin{scope}[shift={(#1)},scale=#2,rotate=#3]
    \Rrule{0,0}{1}{  0}{}
    \Rrule{6,0}{1}{180}{}
    \Rrule{6,0}{1}{  0}{}
    \Rrule{3,3}{1}{270}{}
    \Rrule{6,0}{1}{ 90}{}\end{scope}}

\begin{figure}
\vspace{-10pt} 
\centering
\begin{tikzpicture}[shift={(0,0)},scale=.6,black,thick]
  \tikzstyle{node}=[circle,fill=myaqua,draw=black ,inner sep = 0.33mm, outer sep = 0mm] 
  \begin{scope}[shift={(0,0)},scale=.4]
    \draw[very thick] (-26,0) -- (-28,0);
    \draw (-26.8,-4) node {$\Xi^0$};
    \mynode{-26,0}{5}
    \mynode{-28,0}{5}
    \draw[->,thick] (-24,0)--(-23,0);
    \draw (-19.2,-4) node {$\Xi^1$};
    \Rrule{-21,0}{1}{0}{}
    \draw[->,thick] (-16,0)--(-15,0);
    \draw (-8.2,-4) node {$\Xi^2$};
    \RRrule{-13,0}{1}{0}
    \draw[->,thick] (-2.5,0)--(-1.5,0);
    \draw (13.8,-4) node {$\Xi^3$};
    \RRrule{0,0}{1}{  0}{}
    \RRrule{18,0}{1}{180}{}
    \RRrule{18,0}{1}{  0}{}
    \RRrule{9,9}{1}{270}{}
    \RRrule{18,0}{1}{ 90}{}
  \end{scope} 
\end{tikzpicture}
\caption{\small{An example of evolving theoretical fractal graph~\citep{neroli2024fractal}}}
\label{fig:example}
\vspace{-6pt} 
\end{figure}




To effectively utilize fractal properties, we propose a novel FractalGCL framework in this paper, improving the effectiveness of GCL. 
We start by introducing a novel augmentation strategy, \textbf{renormalisation}, to generate positive views which are structurally similar. 
Therefore, the generated views have the same box dimension, implying strong structural similarity.
To ensure that the graph representations capture not only self-similar structures but also explicitly encode fractal-dimension information, we define a \textbf{fractal-dimension–aware contrastive loss} that steers the encoder to embed graphs in a way that respects their intrinsic fractal geometry.
Empirically, the two components already outperform competing models, yet estimating the fractal dimension introduces additional computational overhead.
Consequently, we cut the cost of box-dimension estimation with a theoretical result that approximates the dimension gap as a \textbf{Gaussian perturbation}, making FractalGCL practical and performant. 
Experiments were conducted on both standard graph classification benchmarks and real-world traffic networks, and the results confirm that FractalGCL surpasses prior methods on most individual benchmarks and attains the best average performance overall, underscoring its effectiveness in both theory and practice.

To sum up, our main contributions include:

\begin{itemize}

\item \textbf{Fractal Geometry Meets GCL.}  
      To the best of our knowledge, we are among the first to inject a mathematically fractal viewpoint into graph representation learning and graph contrastive learning, revealing that a global and scale-free structure, which is often overlooked by prior GCL methods, demonstrates significant potential in learning high-quality graph representations and enhancing performance on downstream tasks.

\item \textbf{Theory-Driven FractalGCL Architecture.}  
      Guided by fractal geometry, we improve the existing GCL methods with a novel framework FractalGCL. It integrates renormalisation-based graph augmentations and a fractal-dimension–aware contrastive loss. Renormalisation contributes to generating better positive and negative pairs, while the novel loss further utilizes the fractal property to optimize graph embeddings.

\item \textbf{From Theory to Implementation.}  
      To make FractalGCL practical, we derive a diameter-controlled Gaussian surrogate for the finite renormalisation discrepancy, avoiding repeated box-counting on renormalised graphs.
      We also design an objective-level fallback mechanism that removes fractal corrections when the box-counting fit is unreliable or the graph diameter is too small.

\item \textbf{Notable Performance Gains.}  
      On MalNet-Tiny, we show that FractalGCL can serve as an effective frozen pretraining tool, improving multiple downstream graph classifiers, and on standard TUDataset benchmarks, FractalGCL also achieves strong graph-level representation performance.
    Furthermore, on additional real-world urban traffic tasks, FractalGCL obtains the best average performance among the compared baselines, demonstrating its effectiveness in practical graph-learning scenarios.

\end{itemize}

\section{Are fractal-like graphs common?}
FractalGCL is motivated by fractal graphs.
This raises a natural question: are fractal graphs actually common?

\begin{figure}[H]
\vspace{-0.4cm}
  \centering
  \begin{minipage}[t]{0.6\linewidth}
    \centering
    \includegraphics[width=\linewidth]{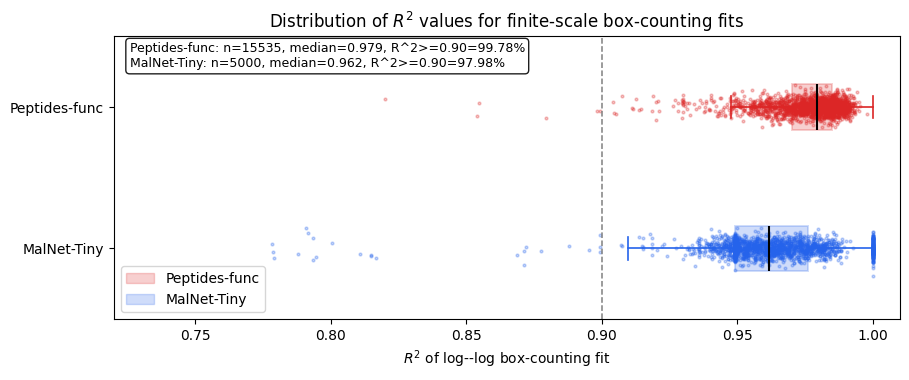}
    \captionof{figure}{\(R^2\) distributions on two large graph benchmarks}
    \label{fig:pre:power}
  \end{minipage}%
  \hfill
  \raisebox{0.5mm}[0pt][0pt]{%
    \begin{minipage}[t]{0.36\linewidth}
      \centering
      \includegraphics[width=\linewidth]{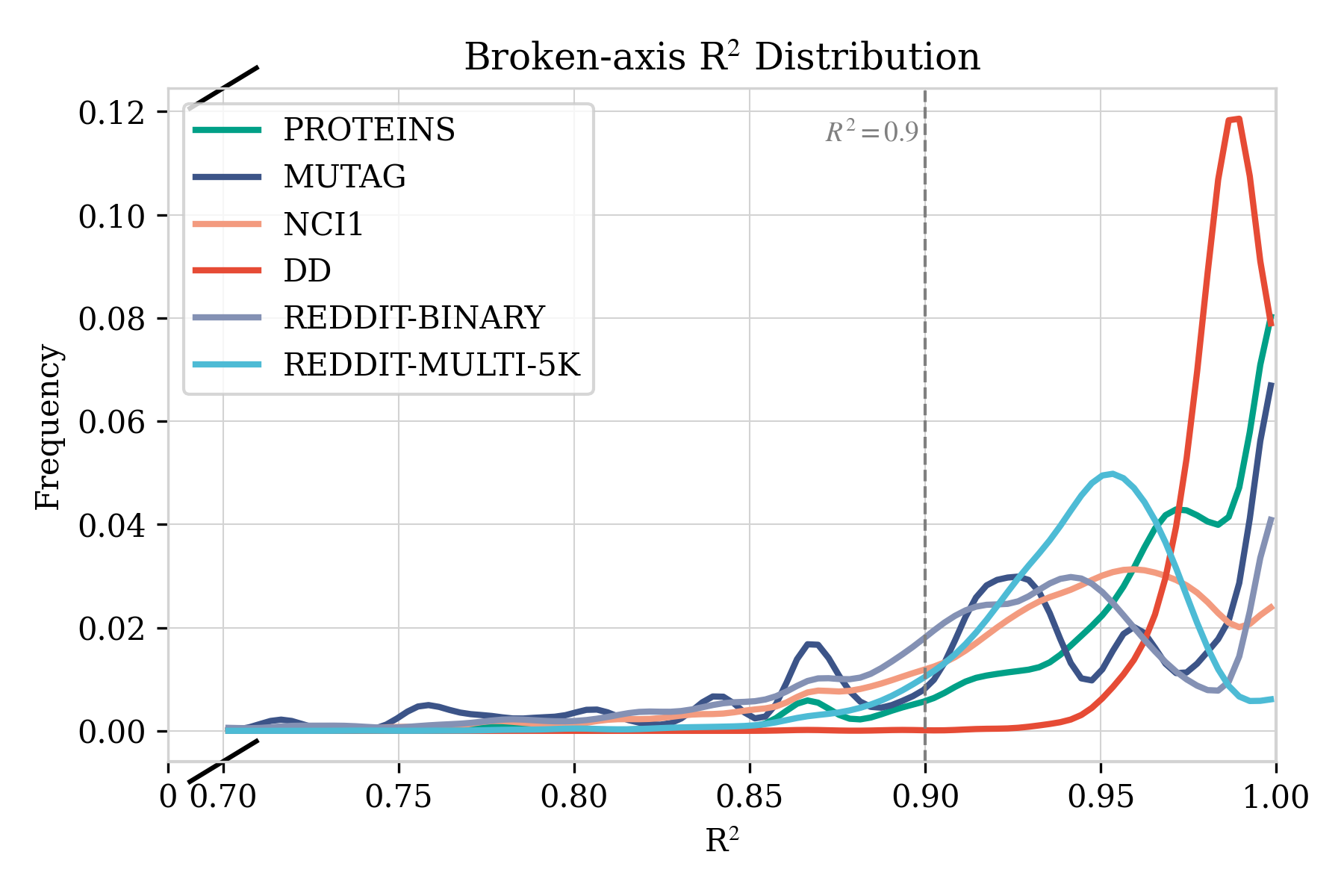}
      \captionof{figure}{\(R^2\) distributions on six standard benchmarks}
      \label{fig:pre:common}
    \end{minipage}%
  }

  \label{fig:pre-exp}
\end{figure}

We use the coefficient of determination \(R^2\) from the log--log box-counting regression as a finite-scale diagnostic of whether benchmark graphs exhibit power-law box-counting behaviour.
On two large graph benchmarks, \textsc{Peptides-func} has a concentrated \(R^2\) distribution with median \(0.979\) and \(99.78\%\) of graphs satisfying \(R^2\ge0.90\), while \textsc{MalNet-Tiny} is more heterogeneous but still strongly regular, with median \(0.962\) and \(97.98\%\) of graphs satisfying \(R^2\ge0.90\); see Figure~\ref{fig:pre:power}.
On standard TUDataset benchmarks, high \(R^2\) values are widespread: using the strict cutoff \(R^2\ge0.90\), \(81\%\) of the \textsc{PROTEINS} graphs, \(92\%\) of the \textsc{REDDIT-MULTI-5K} graphs, and \(99.8\%\) of the \textsc{D\&D} graphs satisfy this criterion; see Figure~\ref{fig:pre:common}.

In fact, fractal behaviour (especially in larger graphs) is very common.
In a corpus of 275 real-world networks, approximately 80\% are classified as fractal~\citep{zakar2023towards}.
For Chinese urban street networks, about 67\% exhibit fractal (power-law) behaviour; moreover, at the level of street connectivities, almost all cities display a fractal hierarchy~\citep{ma2020understanding}.
Examining 47 self-organising networks, 47/47 satisfy the fractal size--density scaling criterion~\citep{laurienti2011universal}.

This naturally raises a further question: given the prevalence of fractal-like graphs, do we have suitable tools for studying them?

\section{FractalGCL: Theory, Methodology and Implementation}

\begin{figure}[ht]
    \vspace{-0.5cm}
    \centering  
    \includegraphics[width=0.85\linewidth]{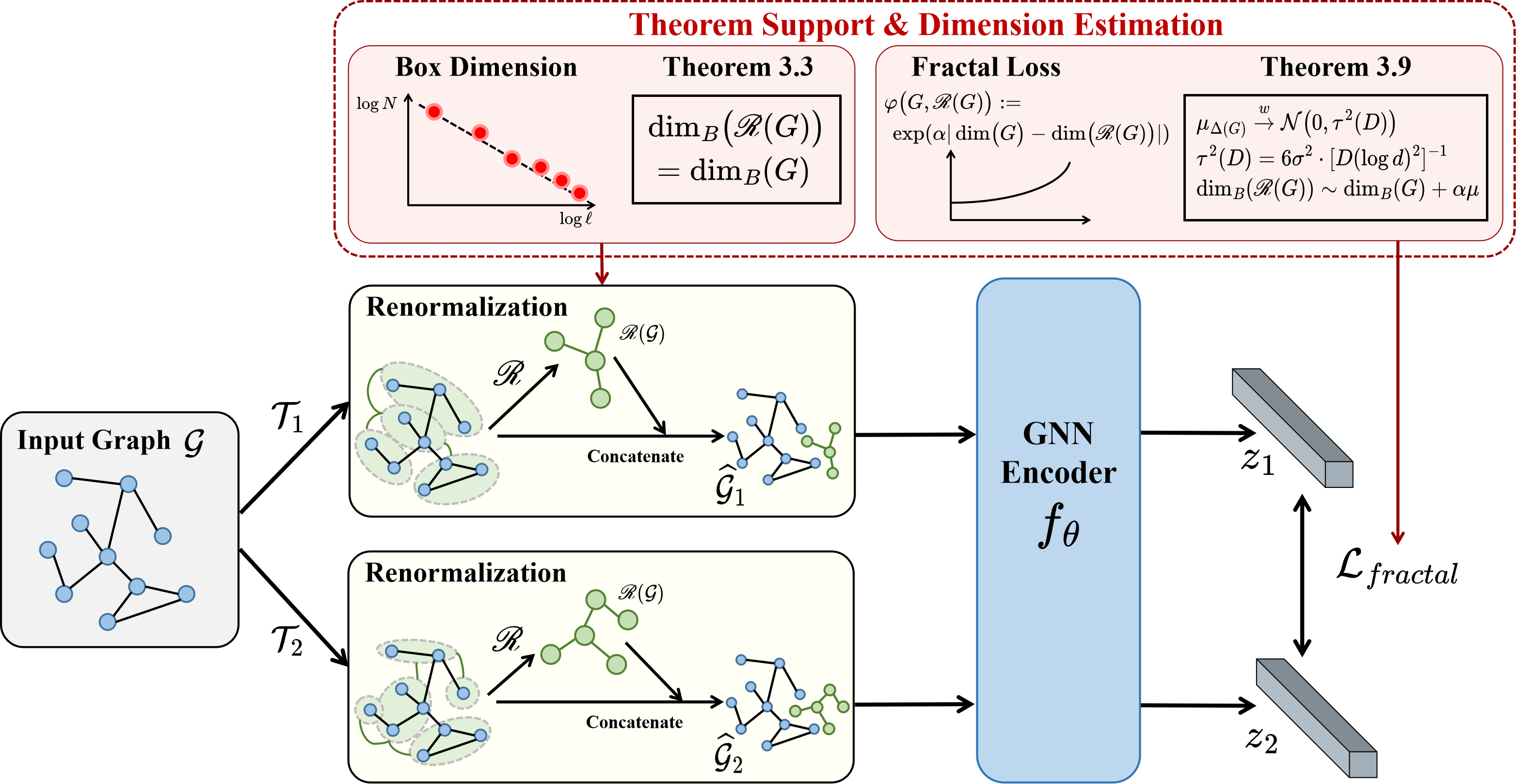}
    \caption{The Pipeline of FractalGCL}
    \label{fig:FRACTALGCL process}
\end{figure}

This section constructs FractalGCL --- a novel framework grounded in fractal geometry that enables graph representations to capture global fractal structure and box dimension information at the graph level. 
Specifically, Section~\ref{subsection:background-of-fractal-learning} revisits the essentials of fractal geometry; Sections~\ref{subsection:augmentation}–\ref{subsection:loss-function} present our renormalisation‑based augmentations and the accompanying dimension‑aware contrastive loss, which form core components of the FractalGCL framework. 
However, computing the fractal loss for each renormalised graph is extremely costly.
Sections~\ref{subsection:dilemma-solution}–\ref{subsection:implementation} address the resulting computational challenge by mathematical proof and statistical analysis and detail the practical implementation of FractalGCL.
See Figure~\ref{fig:FRACTALGCL process} for intuitive ideas.

\subsection{Background of Fractal Geometry}\label{subsection:background-of-fractal-learning}
Mathematically, defining fractals and fractal dimensions is relatively delicate.
From an operational perspective, however, it is enough to view a fractal graph as a graph that approximately satisfies \(N_{\ell}(G)\sim \ell^{-\tdimb(G)}\), where \(\ell\) is a scale parameter and \(N_{\ell}(G)\) denotes the minimum number of balls of radius \(\ell\) needed to cover the graph \(G\).
Here \(\tdimb(G)\) is the estimated Minkowski dimension, also often called the box dimension.
Intuitively, fractality refers to a form of similarity across different scales.

The formal definitions of box partitions, the limiting dimension \(\dimb(G^\infty)\), and the finite-scale estimator \(\tdimb(G)\) are given in Definition~\ref{definition: box_dimension} in Appendix~\ref{appendix: math}.
 When a graph sequence admits a Gromov--Hausdorff scaling limit and the finite-scale box-counting curves converge uniformly, Appendix~\ref{appendix: estimator_consistency} shows that \(\tdimb(G^n)\) converges to the limiting box dimension.

\subsection{New Augmentation: Graph Renormalisation}
\label{subsection:augmentation}

This section introduces the renormalised graph $\mathscr{R}(G)$ used by FractalGCL and states its asymptotic stability property.  The central idea is to replace local groups of vertices by supervertices, producing a coarse graph that retains large-scale structure while removing fine local details.

\textbf{Renormalisation Graph.}

\begin{wrapfigure}[10]{r}{0.5\textwidth}
    \vspace{-1.3cm}
    \includegraphics[width=1\linewidth]{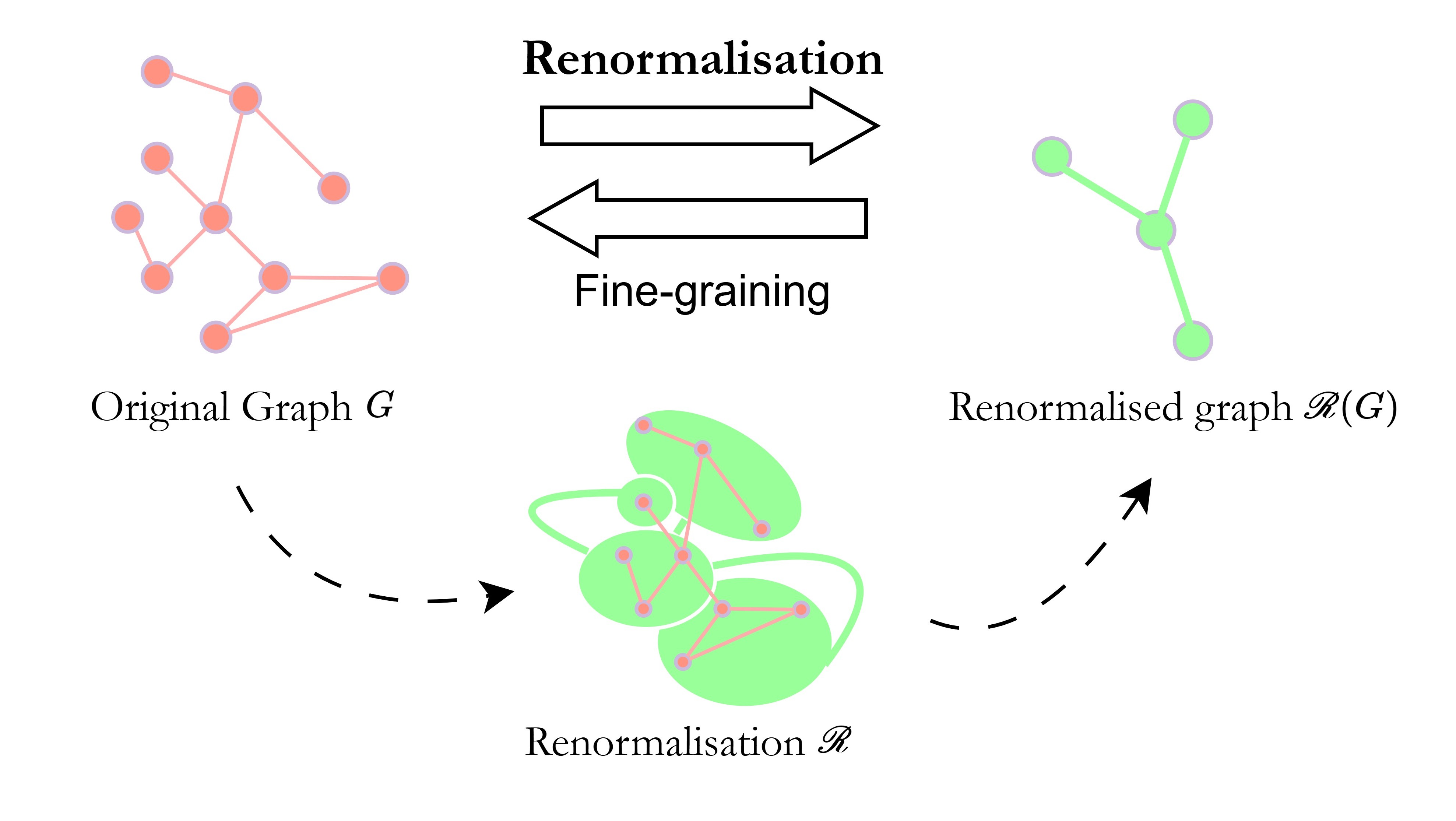}
    \caption{Graph Renormalisation}
    \label{fig:enter-label}
\end{wrapfigure}

In contrastive learning, constructing an augmented graph that remains structurally aligned with the original graph is crucial.  Renormalisation provides such a view by explicitly coarsening the graph through disjoint boxes.  The idea is classical in multi-scale network analysis and complex-network renormalisation~\citep{song2005self}; here we use it as a graph-level augmentation for contrastive learning.

\begin{definition}\label{definition: renormalisation}
Let $G=(V,E)$ be a connected graph and let $\{U_i\}_{i\in\mathcal I}$ be a disjoint partition of $V$ into nonempty connected vertex sets satisfying
\[
V=\bigsqcup_{i\in\mathcal I} U_i,
\qquad
\diam(U_i)\le L
\quad \text{for all } i\in\mathcal I .
\]
The \emph{renormalised graph} $\mathscr{R}(G)$ induced by this partition is defined as follows.  Each box $U_i$ is collapsed into a supervertex $v_i$.  For $i\ne j$, we place a superedge $(v_i,v_j)$ in $\mathscr{R}(G)$ whenever there exists an edge $(u,u')\in E$ with $u\in U_i$ and $u'\in U_j$.  The graph $\mathscr{R}(G)$ is equipped with the unweighted shortest-path metric induced by its adjacency matrix.  For attributed graphs, the feature of a supervertex is obtained by pooling the features of the vertices in its corresponding box.
\end{definition}

The disjointness condition is important: every original vertex is assigned to exactly one supervertex, so the collapse map from $G$ to $\mathscr{R}(G)$ is well-defined.  The greedy procedure in Appendix~\ref{appendix: algorithm_renormalisation} constructs precisely such a disjoint partition by removing vertices once they have been assigned to a box.

\begin{theorem}\label{theorem: renormalisation_keep_dimension}
If the two box dimensions exist under Definition~\ref{definition: box_dimension}, then
\[
\dimb\bigl(\mathscr{R}(G^\infty)\bigr)
=
\dimb(G^\infty).
\]
\end{theorem}
\begin{proof}
    See Theorem~\ref{appendix: theorem_renormalisation_keep_dimension} in Appendix.
\end{proof}

Theorem~\ref{theorem: renormalisation_keep_dimension} is an asymptotic statement about graph sequences.  For a finite graph, the empirical quantities $\tdimb(G)$ and $\tdimb(\mathscr R(G))$ may differ because the regression is performed over finitely many scales.  FractalGCL uses this finite-scale discrepancy as part of the contrastive signal in later sections.

In all FractalGCL augmentations, the augmented view is the disjoint union
$
G\sqcup\mathscr{R}(G).
$
This view gives the encoder access to both the original local structure and a coarse renormalised module.  The box-counting quantities used by the fractal loss are computed on $G$ and $\mathscr{R}(G)$ separately, not on the disconnected union $G\sqcup\mathscr{R}(G)$.  See Figure~\ref{fig:FRACTALGCL process}.

\subsection{Fractal-Dimension-Aware Contrastive Loss}\label{subsection:loss-function}

We now define the contrastive objective used by FractalGCL.  The objective follows the graph-level GCL convention: an anchor graph and its augmented view form a positive pair, while augmented views generated from other graphs in the same minibatch serve as negative samples.  This section presents the one-direction objective used as the theoretical framework; symmetric variants can be introduced later as implementation details.

\textbf{Augmented graph.}
For a graph $G$, define the renormalisation-based augmented graph
\[
\mathcal A(G):=G\sqcup \mathscr R(G),
\]
where $\mathscr R(G)$ is the renormalised quotient graph defined in Section~\ref{subsection:augmentation}.  The disjoint union $\mathcal A(G)$ is the input graph seen by the encoder: it keeps the original graph and appends a coarse-scale renormalised component.  The box-counting quantities used by the loss are not computed on the disconnected union $G\sqcup \mathscr R(G)$; they are computed on the connected original and renormalised components separately.

\textbf{Graph representations.}
For the $i$-th graph $G_i$ in a minibatch, the original graph and its augmented graph are encoded by the same GNN encoder $f_\theta$, readout function, and projection head $g_\phi$:
\[
\mathbf z_i
=
g_\phi\left(
\operatorname{Readout}\left(f_\theta\left(G_i\right)\right)
\right),
\qquad
\mathbf z_i^{(\mathcal A)}
=
g_\phi\left(
\operatorname{Readout}\left(f_\theta\left(\mathcal A(G_i)\right)\right)
\right).
\]
The pair $(\mathbf z_i,\mathbf z_i^{(\mathcal A)})$ is the positive pair for $G_i$.  For $j\neq i$, the pair $(\mathbf z_i,\mathbf z_j^{(\mathcal A)})$ is an in-batch negative pair.

\textbf{GCL-style objective with fractal-aware logits.}
Let $\operatorname{sim}(\cdot,\cdot)$ denote the similarity function used by the contrastive objective, e.g., cosine similarity.  For anchor $G_i$, FractalGCL defines
\[
\ell_i
:=
-\log
\frac{
\exp\left(
\operatorname{sim}\left(
\mathbf z_i,
\mathbf z_i^{(\mathcal A)}
\right)/\tau
-
\alpha
\left|
\tdimb(G_i)
-
\tdimb\left(\mathscr R(G_i)\right)
\right|
\right)
}{
\sum_{\substack{j=1\\j\neq i}}^{N}
\exp\left(
\operatorname{sim}\left(
\mathbf z_i,
\mathbf z_j^{(\mathcal A)}
\right)/\tau
+
\alpha
\left|
\tdimb(G_i)
-
\tdimb\left(\mathscr R(G_j)\right)
\right|
\right)
}.
\]

Following the graph-level GCL convention used in GraphCL, the denominator in
\(\ell_i\) contains only in-batch negatives, while the positive pair appears in
the numerator. We therefore interpret \(\ell_i\) as a log-ratio objective between
the positive logit and the aggregate negative logit, rather than as a normalized
probability over all candidates. If the positive term is also inserted into the
denominator, the corresponding full-softmax loss becomes
\[
\log\left(1+\exp(\ell_i)\right),
\]
which is a strictly increasing transformation of the same log-ratio. Thus the
two objectives encourage the same ordering between the positive logit and the
aggregate negative logit, while inducing different gradient scales. We keep the
negative-only form to remain consistent with the graph-level GCL protocol used
by our baselines. Since \(\operatorname{sim}\) is cosine similarity between
normalized embeddings and the box-counting quantities are finite for every
finite graph, the objective is finite on every finite minibatch.

The total one-direction objective is
$
\mathcal L_{\mathrm{fractal}}
=
\frac{1}{N}
\sum_{i=1}^{N}\ell_i.
$
When $\alpha=0$, the fractal correction vanishes and the objective reduces to the underlying GCL contrastive objective.  When $\alpha>0$, the positive logit is penalised when the renormalised view deviates from the anchor in finite-scale box dimension, while the negative logits become sensitive to the box-dimension discrepancies induced by graph renormalisation.  The quantities involving $\tdimb(\mathscr R(G))$ are the computationally expensive part of the objective; Sections~\ref{subsection:dilemma-solution} and~\ref{subsection:implementation} describe how they are approximated efficiently in practice.

\begin{theorem}
\label{theorem:fractal_negative_repulsion}
For every \(j\neq i\),
\[
\frac{
\partial \ell_i
}{
\partial
\operatorname{sim}\left(
\mathbf z_i,
\mathbf z_j^{(\mathcal A)}
\right)
}
=
\frac{1}{\tau}
\frac{
\exp\left(
\operatorname{sim}\left(
\mathbf z_i,
\mathbf z_j^{(\mathcal A)}
\right)/\tau
+
\alpha
\left|
\widetilde{\dim}_{\mathrm B}(G_i)
-
\widetilde{\dim}_{\mathrm B}\left(\mathscr R(G_j)\right)
\right|
\right)
}{
\sum_{\substack{k=1\\k\neq i}}^{N}
\exp\left(
\operatorname{sim}\left(
\mathbf z_i,
\mathbf z_k^{(\mathcal A)}
\right)/\tau
+
\alpha
\left|
\widetilde{\dim}_{\mathrm B}(G_i)
-
\widetilde{\dim}_{\mathrm B}\left(\mathscr R(G_k)\right)
\right|
\right)
}.
\]
\end{theorem}

\begin{proof}
    See Appendix~\ref{appendix:theorem_fractal_negative_repulsion}.
\end{proof}

The theorem shows that negatives with larger finite-scale box-counting discrepancy receive stronger repulsion, while box-counting-similar graphs remain negatives but are repelled less strongly.
Also see~\ref{corollary:larger_discrepancy_stronger_repulsion}, Corollary~\ref{corollary:dimension_similar_not_positive}  and Corollary~\ref{corollary:dimension_dependent_margin} in Appendix \ref{appendix: math}.
This dimension-aware term is a structural prior, not a label-semantic oracle: it does not assume that graph labels are determined only by box-counting dimension.
Instead, it biases the standard in-batch negative objective toward separating graphs with substantially different finite-scale geometry.
The strength of this bias is controlled by \(\alpha\), and Section~\ref{subsection:safe-fallback} disables the fractal correction when the box-counting fit is weak or the graph diameter is too small.

\subsection{Computational Dilemma and Its Solution}
\label{subsection:dilemma-solution}

The loss in Section~\ref{subsection:loss-function} involves $\tdimb(\mathscr R(G))$, the finite-scale box-counting dimension of a renormalised graph.  Directly recomputing this quantity for every augmented graph is expensive, because each computation requires a multi-scale box-covering regression.  This section analyzes the bottleneck and introduces a diameter-controlled Gaussian surrogate for the finite renormalisation error.

To summarize, FractalGCL faces the following dilemma.

\textbf{(I)} By Theorem~\ref{theorem: renormalisation_keep_dimension}, bounded renormalisation preserves box dimension in the infinite-diameter limit.  However, for a finite graph, the empirical discrepancy
$
\Delta(G)
:=
\tdimb(G)-\tdimb(\mathscr R(G))
$
need not vanish exactly.

\textbf{(II)} Computing $\tdimb(\mathscr R(G))$ exactly for every augmented graph is computationally expensive.  The following proposition states the resulting box-counting cost.

\begin{proposition}
\label{proposition: fractal_complexity}
Under a cached-neighbourhood implementation of Algorithm~1, all-scale greedy box covering costs
\[
O\!\left(|\mathcal L(G)|\,|V|(|V|+|E|)\right).
\]
For sparse graphs with $|E|=O(|V|)$ and $|\mathcal L(G)|=O(|V|)$, this gives $O(|V|^3)$, and path-like graphs require $\Omega(|V|^2)$ operations.
\end{proposition}
\begin{proof}
    See Appendix~\ref{appendix: proposition_fractal_complexity}.
\end{proof}

\textbf{A stochastic solution to the dimension dilemma.}
Instead of recomputing $\tdimb(\mathscr R(G))$, we approximate the finite renormalisation discrepancy $\Delta(G)$ by a Gaussian surrogate
\[
\widetilde{\Delta}(G)
\sim
\mathcal N\!\left(
0,
\sigma^2\!\left(\widetilde{\Delta}(G)\right)
\right).
\]
The surrogate variance is controlled by the uncertainty of the finite-scale box-counting regression.
We defer the detailed construction of this surrogate to Appendix~\ref{appendix: math}.
After Proposition~\ref{proposition: fractal_complexity} identifies the computational bottleneck, Lemma~\ref{appendix: lemma:_diameter_infinity} gives a diameter-controlled standard-error scale, Lemma~\ref{appendix: lemma_ols_clt} establishes the normal fluctuation of the finite-scale estimator, and Proposition~\ref{proposition: gaussian_surrogate} combines these ingredients into the Gaussian surrogate used below.

\begin{proposition}
\label{proposition: gaussian_surrogate}
Under the finite-scale regression fluctuation model in Lemma~\ref{appendix: lemma_ols_clt}, and with the variance scale calibrated by \(\lambda_{\Delta}\), the centred renormalisation discrepancy admits the Gaussian approximation
\[
\frac{
\Delta(G)-\mathbb E[\Delta(G)]
}{
\sigma\!\left(\widetilde{\Delta}(G)\right)
}
\xrightarrow{\mathscr D}
\mathcal N(0,1)
\qquad
\text{as } \diam(G)\to\infty .
\]
\end{proposition}
\begin{proof}
    See Appendix~\ref{appendix: proposition_gaussian_surrogate}.
\end{proof}

The proposition should be read as a conditional surrogate result, not as a distribution-free finite-sample law. It is motivated by asymptotic renormalisation stability and by a finite-scale regression fluctuation model; the variance inflation factor absorbs the additional uncertainty and dependence introduced by the renormalised graph.

\noindent\textbf{Summary.}
Since
$
\Delta(G)
=
\tdimb(G)-\tdimb(\mathscr R(G)),
$
we approximate
$
\tdimb(\mathscr R(G))
\approx
\tdimb(G)-\widetilde{\Delta}(G),
$
where
$
\widetilde{\Delta}(G)
\sim
\mathcal N\!\left(
0,
\sigma^2\!\left(\widetilde{\Delta}(G)\right)
\right).
$

\subsection{Practical Implementation}
\label{subsection:implementation}

We now describe how the surrogate in Section~\ref{subsection:dilemma-solution} is used during training.  The implementation avoids recomputing $\tdimb(\mathscr R(G))$ for each augmented graph.


\textbf{Sampling surrogate renormalisation errors.}
For a minibatch $\{G_1,\ldots,G_N\}$, we sample one surrogate discrepancy per graph:
$
\widetilde{\Delta}(G_i)
\sim
\mathcal N\!\left(
0,
\sigma^2\!\left(\widetilde{\Delta}(G_i)\right)
\right),$
for $i=1,\ldots,N.$
The same $\widetilde{\Delta}(G_i)$ is reused in all pairwise logits involving the renormalised dimension of $G_i$.  This reflects the approximation
$
\tdimb(\mathscr R(G_i))
\approx
\tdimb(G_i)-\widetilde{\Delta}(G_i).
$

\textbf{Fractal-corrected logits.}
Let
\[
\mathbf z_i
=
g_\phi\!\left(
\operatorname{Readout}(f_\theta(G_i))
\right),
\qquad
\mathbf z_i^{(\mathcal A)}
=
g_\phi\!\left(
\operatorname{Readout}(f_\theta(\mathcal A(G_i)))
\right).
\]
We form the corrected logit matrix
\[
\mathbf S^\ast_{ij}
=
\frac{
\operatorname{sim}
\left(
\mathbf z_i,\mathbf z_j^{(\mathcal A)}
\right)
}{\tau}
+
\alpha \mathbf C_{ij},
\]
where
\[
\mathbf C_{ij}
=
\begin{cases}
-\left|\widetilde{\Delta}(G_i)\right|,
& i=j,\\[4pt]
\left|
\tdimb(G_i)
-
\tdimb(G_j)
+
\widetilde{\Delta}(G_j)
\right|,
& i\neq j.
\end{cases}
\]
The diagonal correction corresponds to the positive-view reliability penalty, whereas the off-diagonal correction approximates the negative discrepancy
$
\left|
\tdimb(G_i)-\tdimb(\mathscr R(G_j))
\right|.
$

The one-direction GCL-style minibatch loss is
$
\ell_i
=
-\log
\frac{
\exp(\mathbf S^\ast_{ii})
}{
\sum_{\substack{j=1\\j\neq i}}^N
\exp(\mathbf S^\ast_{ij})
},
$
and
$
\mathcal L_{\mathrm{fractal}}
=
\frac1N\sum_{i=1}^N\ell_i.
$
This adds only $O(N^2)$ arithmetic per minibatch after the graph-level box-counting statistics have been cached.

\subsection{Safe fallback under weak fractality or small diameters}
\label{subsection:safe-fallback}
We employ a gate based on finite-scale reliability.  If $\diam(G)\le9$, or if the box-counting fit is weak with $R^2(G)<\theta$ for default $\theta=0.9$, we set $\alpha=0$ and disable the fractal correction.
With $\alpha=0$, all fractal-dimension terms in Section~\ref{subsection:loss-function} vanish, and the loss takes the standard GCL-style form with a positive pair in the numerator and only in-batch negatives in the denominator.
Thus, in weak-fractality or small-diameter regimes, FractalGCL falls back at the objective level to the underlying GCL loss form and does not impose additional fractal bias.  This is an objective-level safeguard rather than a deterministic guarantee on downstream accuracy.
See Section~\ref{subsection:parameter} for parameter analysis.

\section{Experiments}

\subsection{Validation of the Gaussian surrogate}

We aim to test that the change in the Minkowski dimension after one-step renormalisation $\Delta(G)$ is Gaussian in distribution.
On the TUDataset D\&D benchmark, we find $\mathrm{mean}(\Delta(G))=-0.1084$ and $\mathrm{std}=0.1058$ with $n=1178$; 
the regression $D'=\alpha+\beta D$ yields $\alpha=0.0512$, $\beta=0.9711$ and $R^2=0.87$, and $\mathrm{corr}(D, D'-D)=0.0214$ $(p=0.46)$.
Hence, the experimental evidence supports the Gaussian surrogate in Proposition~\ref{proposition: gaussian_surrogate}.

\subsection{Frozen pretraining experiment on {MalNet}}

In this experiment, we aim to demonstrate the ability of FractalGCL to serve as a pretraining tool.

We evaluate FractalGCL as a frozen self-supervised pretraining signal on \textsc{MalNet-Tiny}, rather than as a standalone fully supervised predictor.
The FractalGCL encoder is pretrained without malware labels, frozen, and fused with three downstream graph neural network heads.
We use a three-layer GIN pretraining encoder with hidden dimension 256 and max pooling, pretrain it for 15 epochs with \(\alpha\) linearly decayed from \(0.05\) to \(0.01\), and report averages over three seeds.
All runs were conducted on a single NVIDIA A100-SXM4-40GB GPU with CUDA 12.8 and PyTorch 2.10.0+cu128.

\begin{figure*}[h]
\centering

\begin{minipage}[t]{0.26\textwidth}
\centering
\vspace{0pt}
\captionsetup{font=footnotesize}
\captionof{figure}{Gaussian validation}
\label{fig:gaussian_validation}
\vspace{0.15cm}
\includegraphics[width=\linewidth]{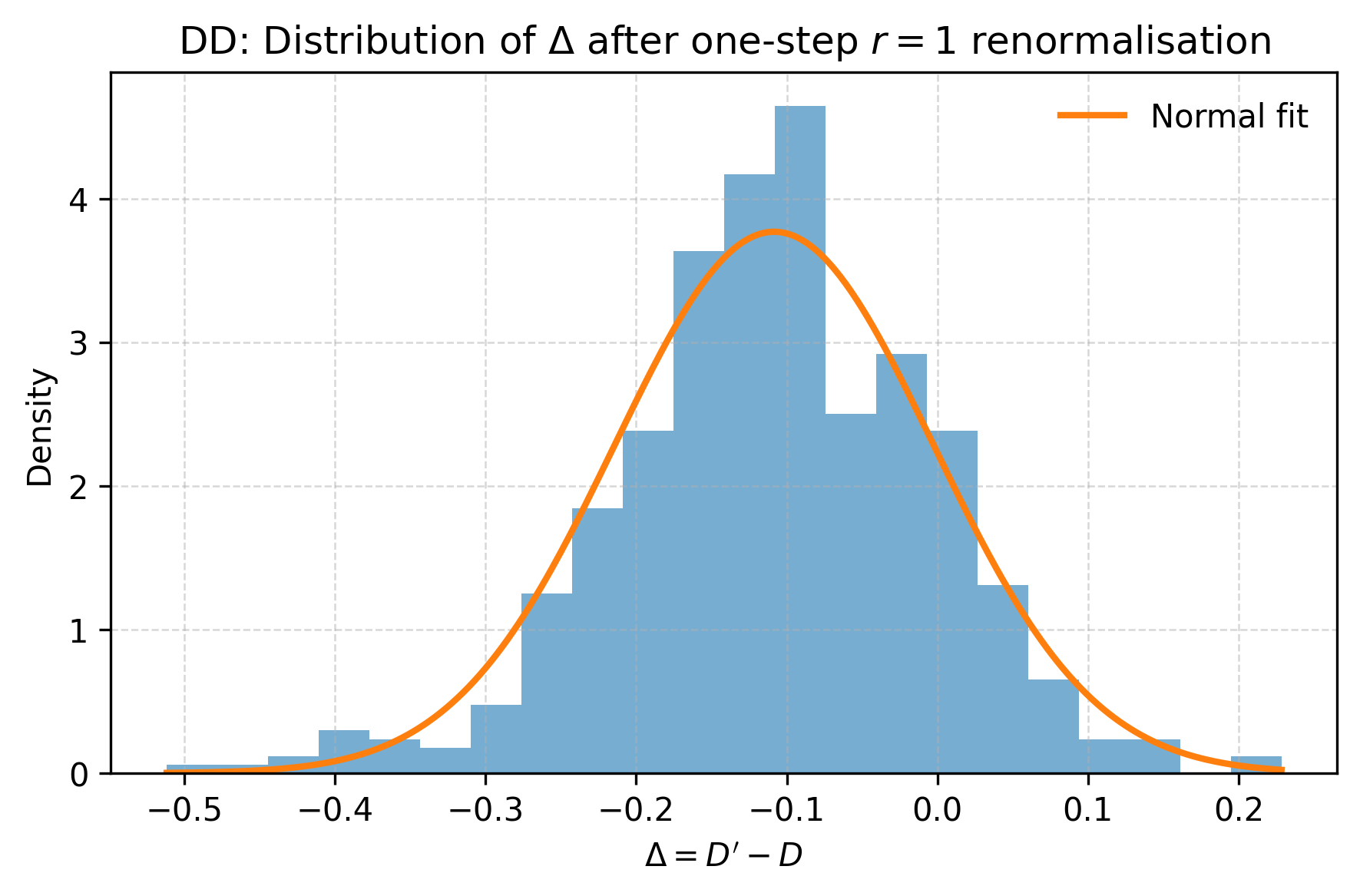}
\end{minipage}
\hfill
\begin{minipage}[t]{0.70\textwidth}
\centering
\vspace{0pt}
\captionof{table}{Frozen pretraining results on \textsc{MalNet-Tiny}.}
\label{tab:frozen_pretraining}
\renewcommand{\arraystretch}{0.84}
\tiny

\begin{subtable}[t]{0.53\linewidth}
\centering
\caption{Published results.}
\label{tab:malnet_published}
\resizebox{\linewidth}{!}{%
\begin{tabular}{lc}
\toprule
\textbf{Model} & \textbf{Accuracy (\%)} \\
\midrule
GCN-JK~\citep{lo2022android}                & 89.70 \\
GIN-JK~\citep{lo2022android}                & 90.00 \\
GraphSAGE~\citep{cao2023gst}                & 88.08 $\pm$ 1.68 \\
GST-SAGE~\citep{cao2023gst}                 & 88.42 $\pm$ 1.03 \\
Graph-Mamba~\citep{wang2024graph}           & 93.40 $\pm$ 0.27 \\
GCN+~\citep{luo2025can}                     & 93.54 $\pm$ 0.45 \\
GIN+~\citep{luo2025can}                     & 93.25 $\pm$ 0.40 \\
\bottomrule
\end{tabular}%
}
\end{subtable}
\hfill
\begin{subtable}[t]{0.42\linewidth}
\centering
\caption{Our runs.}
\label{tab:malnet_ours}
\resizebox{\linewidth}{!}{%
\begin{tabular}{lc}
\toprule
\textbf{Model} & \textbf{Accuracy (\%)} \\
\midrule
Frozen-FractalGCL        & 85.13 $\pm$ 0.61 \\
Random frozen control    & 18.83 $\pm$ 1.03 \\
GIN-small                & 90.10 $\pm$ 0.96 \\
FractalGCL + GIN-small   & \textbf{90.33 $\pm$ 0.76} \\
SAGE-small               & 90.93 $\pm$ 0.61 \\
FractalGCL + SAGE-small  & \textbf{92.13 $\pm$ 0.29} \\
JK-SAGE                  & 92.93 $\pm$ 0.15 \\
FractalGCL + JK-SAGE     & \textbf{93.40 $\pm$ 0.36} \\
\bottomrule
\end{tabular}%
}
\end{subtable}

\end{minipage}
\end{figure*}

The frozen FractalGCL branch improves all three matched downstream classifiers, with the largest gain of \(1.20\) percentage points for \textsc{SAGE-small}.

\subsection{Experiment on TUDataset}

We validate FractalGCL on unsupervised representation learning tasks using five widely-adopted datasets from TUDataset \citep{morris2020tudataset}: \textsc{NCI1}, \textsc{MUTAG}, \textsc{PROTEINS}, \textsc{D\&D}, and \textsc{REDDIT-MULTI-5K} (REDDIT-M5K).
We adopt a 2-layer GIN as the encoder, and a sum pooling is used as the readout function;
renormalisation adopts greedy box-covering with radius~1, dimension weight $\alpha=0.1$, and temperature $\tau=0.4$.  
Models are first trained with Adam on the unlabeled data only.
After that, a non-linear SVM classifier is used to evaluate the graph representations.  
Accuracy is reported under 10-fold cross-validation. The experiments are repeated 5 times to report the mean and standard deviation. 
We conduct our experiments on an Ubuntu machine with one 40GB NVIDIA A100 GPU.


\subsubsection{Main Results}
\begin{table}[ht]
\vspace{-0.3cm}
\centering
\caption{Classification accuracy on benchmark datasets (10-fold CV).}
\label{tab:main_results}
\resizebox{\linewidth}{!}{%
\begin{tabular}{lcccccc}
\toprule
\textbf{Model} & \textbf{NCI1} & \textbf{MUTAG} & \textbf{PROTEINS} & \textbf{D\&D} & \textbf{REDDIT-M5K} & \textbf{AVG.}\\
\midrule
GAE~\citep{kipf2016variational}       & 74.36 $\pm$ 0.24 & 72.87 $\pm$ 6.84    & 70.51 $\pm$ 0.17 & 74.54 $\pm$ 0.68 & 33.58 $\pm$ 0.13     & 65.17 $\pm$ 1.61 \\
graph2vec~\citep{narayanan2017graph2vec}   & 73.22 $\pm$ 1.81 & 83.15 $\pm$ 9.25 & 73.30  $\pm$ 2.05 & 70.32 $\pm$ 2.32 & 47.86 $\pm$ 0.26 & 69.57 $\pm$ 3.14 \\
DGI~\citep{velickovic2019deep} & 74.86 $\pm$ 0.26 & 66.49 $\pm$ 2.28    & 72.27 $\pm$ 0.40 & 75.78 $\pm$ 0.34 & 53.61 $\pm$ 0.31     & 68.60 $\pm$ 0.72 \\
InfoGraph~\citep{sun2019infograph}   & 76.20 $\pm$ 1.06  & 89.01 $\pm$ 1.13 & 74.44 $\pm$ 0.31 & 72.85 $\pm$ 1.78 & 53.46 $\pm$ 1.03 & 73.19 $\pm$ 1.06 \\
GraphCL~\citep{you2020graph}     & 77.87 $\pm$ 0.41 & 86.80 $\pm$ 1.34 & 74.39 $\pm$ 0.45 & 78.62 $\pm$ 0.40 & 55.99 $\pm$ 0.28 & 74.73 $\pm$ 0.58 \\
ContextPred~\citep{hu2020strategies} & 73.00 $\pm$ 0.30 & 71.75 $\pm$ 7.34    & 70.23 $\pm$ 0.63 & 74.66 $\pm$ 0.51 & 51.23 $\pm$ 0.84     & 68.17 $\pm$ 1.92 \\
JOAO~\citep{you2021graph}        & 78.07 $\pm$ 0.47 & 87.35 $\pm$ 1.02 & 74.55  $\pm$ 0.41 & 77.32 $\pm$ 0.54 & 55.74 $\pm$ 0.63 & 74.61 $\pm$ 0.61 \\
JOAOv2~\citep{you2021graph}     & 78.36 $\pm$ 0.53 & 87.67 $\pm$ 0.79 & 74.07 $\pm$ 1.10 & 77.40 $\pm$ 1.15 & 56.03 $\pm$ 0.27 & 74.71 $\pm$ 0.77 \\
SimGRACE~\citep{xia2022simgrace}    & {79.12 $\pm$ 0.44} & {89.01 $\pm$ 1.31} & 74.03 $\pm$ 0.09 & 77.44 $\pm$ 1.11 & 55.91 $\pm$ 0.34 & 75.10 $\pm$ 0.66 \\
RGCL~\citep{li2022let}        & 78.14 $\pm$ 1.08 & 87.66 $\pm$ 1.01 & 75.03 $\pm$ 0.43 & 78.86 $\pm$ 0.48 & \underline{56.38 $\pm$ 0.40} & 75.21 $\pm$ 0.68 \\
DRGCL~\citep{ji2024rethinking} & 78.70 $\pm$ 0.40 & \underline{89.50 $\pm$ 0.60} & \underline{75.20 $\pm$ 0.60} & 78.40 $\pm$ 0.70 & 56.30 $\pm$ 0.20 & 75.62 $\pm$ 0.50 \\
GradGCL~\citep{li2024gradgcl} & \underline{79.72 $\pm$ 0.53} & 88.46 $\pm$ 0.98 & 74.89 $\pm$ 0.39 & \underline{78.95 $\pm$ 0.47} & 56.20 $\pm$ 0.31 & \underline{75.64 $\pm$ 0.54} \\
MGCL~\citep{azizpour2025model}
& 78.66 $\pm$ 0.34
& 89.55 $\pm$ 1.08
& 74.85 $\pm$ 0.71
& 78.88 $\pm$ 0.38
& 55.65 $\pm$ 0.32
& 75.52 $\pm$ 0.57 \\
\midrule
FractalGCL  & \textbf{80.50 $\pm$ 0.16} & \textbf{91.71 $\pm$ 0.23} & \textbf{75.85 $\pm$ 0.40} & \textbf{80.14 $\pm$ 0.12} &  \textbf{56.45 $\pm$ 0.14}  & \textbf{76.93 $\pm$ 0.21} \\
\bottomrule
\end{tabular}%
}
\end{table}

Table~\ref{tab:main_results} reports the accuracy of the downstream graph classification task on five
benchmark datasets.
FractalGCL achieves the highest average score
(\textbf{76.93}\%), outperforming the strongest baseline
(GradGCL, $75.64\%$) by \textbf{1.29 pp}. 
It ranks first on all five retained datasets—\textbf{NCI1}, \textbf{MUTAG},
\textbf{PROTEINS}, \textbf{D\&D}, and \textbf{REDDIT-MULTI-5K}. 
The most strongly fractal benchmark D\&D exhibits a margin of \textbf{1.19 pp}, 
which is consistent with our hypothesis that fractal-aware augmentations and loss provide greater benefit when the underlying graphs display pronounced self-similarity.
These results confirm that injecting fractal structure into graph contrastive learning not only matches but often exceeds the performance of carefully tuned augmentation-based methods, while retaining the same encoder capacity and training budget.

\subsubsection{Ablation Study}

\begin{table}[ht]
\vspace{-0.3cm}
\centering
\begin{minipage}[t]{0.5\linewidth}
  \centering
  \caption{\textbf{(a) Ablation study}}
  \label{tab:ablation}
  \resizebox{0.92\linewidth}{!}{
  \begin{tabular}{lcccc}
    \toprule
    \multirow{2}{*}{Method} &
      \multicolumn{2}{c}{Components} &
      \multicolumn{2}{c}{MUTAG} \\
    \cmidrule(lr){2-3}\cmidrule(lr){4-5}
      & Ren. & Frac. Loss & Acc. & Time (s) \\
    \midrule
    FractalGCL & \checkmark & \checkmark & \textbf{91.71} & 486.81 \\
    w/o Graph Concat & \checkmark & \checkmark & 90.41 & 321.87 \\
    w/o Renormalisation   & $\times$   & \checkmark & 88.46 & 33.93  \\
    w/o Fractal Loss & \checkmark & $\times$   & 88.09 & 423.97 \\
    w/ Exact Dimension & \checkmark & \checkmark & 90.32 & 1249.74 \\
    \bottomrule
  \end{tabular}}
\end{minipage}
\hfill
\begin{minipage}[t]{0.48\linewidth}
  \centering
  \caption{\textbf{(b) Variant accuracy}}
  \label{tab:variant}
  \resizebox{0.8\linewidth}{!}{
  \begin{tabular}{lcc}
    \toprule
    Variant & D\&D & MUTAG \\
    \midrule
    FractalGCL          & \textbf{80.14} & \textbf{91.71} \\
    + random radius     & 78.78 & 88.73 \\
    + $R^{2}$ prob.     & \underline{79.80} & \underline{88.83}\\
    – $R^{2}$ threshold & 79.63 & 88.33 \\
    \bottomrule
  \end{tabular}}
\end{minipage}
\end{table}

Table \ref{tab:ablation} lists MUTAG accuracy and pre-training time as we remove FractalGCL’s three key components—graph concatenation, renormalisation, and the fractal-dimension loss—one at a time.
Dropping any single component lowers accuracy by about 1.3–3.6 pp, confirming that each part is essential.
In terms of efficiency, our Gaussian surrogate for box-dimension estimation trims training time from 1249.74 s (w/ Exact Dimension) to 486.81 s, nearly a 2.56 × speed-up—that is, roughly a 61\% reduction in compute.

\subsubsection{Variant Experiments}

Table \ref{tab:variant} compares three ways of altering the renormalisation rule.
Introducing a random radius or discarding the fractality filter both weaken the structural match between views and lower accuracy, while using $R^{2}$ merely as a soft sampling probability yields a middle-ground result.
These variants confirm that a fixed small radius combined with an explicit $R^{2}$ threshold offers the best balance between view diversity and global consistency.

\subsubsection{Parameter Analysis}\label{subsection:parameter}

\begin{wrapfigure}[7]{r}{0.55\linewidth}
    \vspace{-1cm}
    \centering
    \includegraphics[width=1\linewidth]{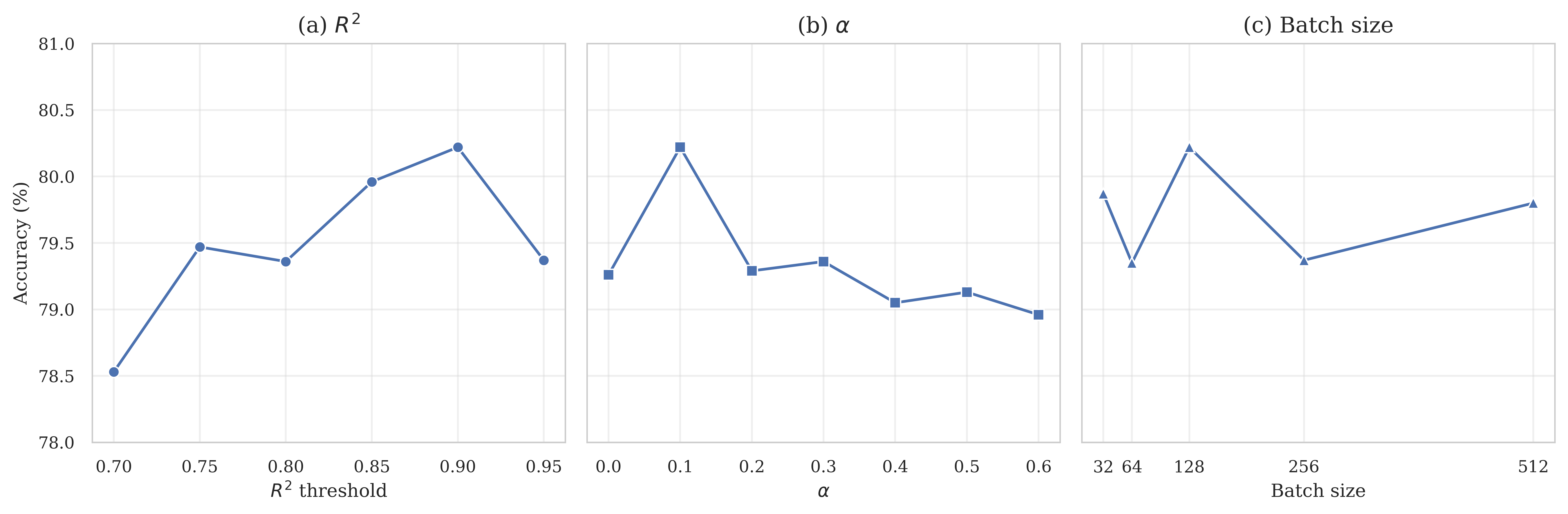}
    \caption{Hyper-parameter sensitivity on D\&D}
    \label{fig:hp}
\end{wrapfigure}

Figure~\ref{fig:hp} shows that FractalGCL remains competitive under reasonable hyper-parameter changes.
Accuracy varies by a few percentage points across the tested settings, with the best performance obtained near $R^2=0.9$.

\subsection{Evaluation on Urban Traffic Networks}
\label{subsection:city}

To test whether FractalGCL transfers beyond curated graph benchmarks, we evaluate frozen graph embeddings on six traffic-safety classification tasks constructed from urban road-network subgraphs in Chicago, San Francisco, and New York.
Table~\ref{tab:city_average} reports the results averaged over all cities and tasks, and the full city-level results and experimental details are provided in Appendix Table~\ref{tab:traffic_tasks_pct}.

\begin{table}[ht]
\vspace{-0.3cm}
\centering
\caption{Average classification accuracy on urban traffic-safety tasks.}
\label{tab:city_average}
\resizebox{\linewidth}{!}{%
\begin{tabular}{lcccccc}
\toprule
\textbf{Model} & DGI & InfoGraph & GCL & JOAO & SimGRACE & \textbf{FractalGCL} \\
\midrule
\textbf{Average accuracy (\%)} 
& 56.12 $\pm$ 11.26
& 56.07 $\pm$ 11.20
& 57.25 $\pm$ 12.57
& 56.04 $\pm$ 12.25
& 56.64 $\pm$ 12.61
& \textbf{61.76 $\pm$ 13.40} \\
\bottomrule
\end{tabular}%
}
\vspace{-0.2cm}
\end{table}

\section{Related Works and Conclusions}
See Appendix \ref{appendix:related} for Related Works and Appendix \ref{sec:conclusion} for Conclusions.

\bibliographystyle{plainnat}
\bibliography{references}

@inproceedings{hu2020strategies,
  title     = {Strategies for Pre-training Graph Neural Networks},
  author    = {Hu, Weihua and Liu, Bowen and Gomes, Joseph and Zitnik, Marinka and Liang, Percy and Pande, Vijay and Leskovec, Jure},
  booktitle = {International Conference on Learning Representations},
  year      = {2020}
}

@inproceedings{xia2022simgrace,
  title     = {{SimGRACE}: A Simple Framework for Graph Contrastive Learning without Data Augmentation},
  author    = {Xia, Jun and Wu, Lirong and Chen, Jintao and Hu, Bozhen and Li, Stan Z.},
  booktitle = {Proceedings of the ACM Web Conference 2022},
  pages     = {1070--1079},
  year      = {2022}
}

@inproceedings{you2021graph,
  title     = {Graph Contrastive Learning Automated},
  author    = {You, Yuning and Chen, Tianlong and Shen, Yang and Wang, Zhangyang},
  booktitle = {International Conference on Machine Learning},
  pages     = {12121--12132},
  year      = {2021},
  publisher = {PMLR}
}

@article{ju2024towards,
  title   = {Towards Graph Contrastive Learning: A Survey and Beyond},
  author  = {Ju, Wei and Wang, Yifan and Qin, Yifang and Mao, Zhengyang and Xiao, Zhiping and Luo, Junyu and Yang, Junwei and Gu, Yiyang and Wang, Dongjie and Long, Qingqing and others},
  journal = {arXiv preprint arXiv:2405.11868},
  year    = {2024}
}

@article{liu2022graph,
  title     = {Graph Self-Supervised Learning: A Survey},
  author    = {Liu, Yixin and Jin, Ming and Pan, Shirui and Zhou, Chuan and Zheng, Yu and Xia, Feng and Yu, Philip S.},
  journal   = {IEEE Transactions on Knowledge and Data Engineering},
  volume    = {35},
  number    = {6},
  pages     = {5879--5900},
  year      = {2022},
  publisher = {IEEE}
}

@article{xu2018powerful,
  title   = {How Powerful are Graph Neural Networks?},
  author  = {Xu, Keyulu and Hu, Weihua and Leskovec, Jure and Jegelka, Stefanie},
  journal = {arXiv preprint arXiv:1810.00826},
  year    = {2018}
}

@article{wu2021self,
  title     = {Self-Supervised Learning on Graphs: Contrastive, Generative, or Predictive},
  author    = {Wu, Lirong and Lin, Haitao and Tan, Cheng and Gao, Zhangyang and Li, Stan Z.},
  journal   = {IEEE Transactions on Knowledge and Data Engineering},
  volume    = {35},
  number    = {4},
  pages     = {4216--4235},
  year      = {2021},
  publisher = {IEEE}
}

@article{xie2022self,
  title     = {Self-Supervised Learning of Graph Neural Networks: A Unified Review},
  author    = {Xie, Yaochen and Xu, Zhao and Zhang, Jingtun and Wang, Zhengyang and Ji, Shuiwang},
  journal   = {IEEE Transactions on Pattern Analysis and Machine Intelligence},
  volume    = {45},
  number    = {2},
  pages     = {2412--2429},
  year      = {2022},
  publisher = {IEEE}
}

@article{chen2025data,
  title     = {Data Scarcity in Recommendation Systems: A Survey},
  author    = {Chen, Zefeng and Gan, Wensheng and Wu, Jiayang and Hu, Kaixia and Lin, Hong},
  journal   = {ACM Transactions on Recommender Systems},
  volume    = {3},
  number    = {3},
  pages     = {1--31},
  year      = {2025},
  publisher = {ACM}
}

@article{shi2025domain,
  title   = {Domain Adaptation for Graph Representation Learning: Challenges, Progress, and Prospects},
  author  = {Shi, Bo-Shen and Wang, Yong-Qing and Guo, Fang-Da and Xu, Bing-Bing and Shen, Hua-Wei and Cheng, Xue-Qi},
  journal = {Journal of Computer Science and Technology},
  pages   = {1--18},
  year    = {2025}
}

@article{liu2023towards,
  title   = {Towards Graph Foundation Models: A Survey and Beyond},
  author  = {Liu, Jiawei and Yang, Cheng and Lu, Zhiyuan and Chen, Junze and Li, Yibo and Zhang, Mengmei and Bai, Ting and Fang, Yuan and Sun, Lichao and Yu, Philip S. and others},
  journal = {arXiv preprint arXiv:2310.11829},
  year    = {2023}
}

@inproceedings{huang2024large,
  title     = {Large Language Models for Graphs: Progresses and Directions},
  author    = {Huang, Chao and Ren, Xubin and Tang, Jiabin and Yin, Dawei and Chawla, Nitesh},
  booktitle = {Companion Proceedings of the ACM Web Conference 2024},
  pages     = {1284--1287},
  year      = {2024}
}

@inproceedings{liu2022towards,
  title     = {Towards Unsupervised Deep Graph Structure Learning},
  author    = {Liu, Yixin and Zheng, Yu and Zhang, Daokun and Chen, Hongxu and Peng, Hao and Pan, Shirui},
  booktitle = {Proceedings of the ACM Web Conference 2022},
  pages     = {1392--1403},
  year      = {2022}
}

@article{rong2019dropedge,
  title   = {{DropEdge}: Towards Deep Graph Convolutional Networks on Node Classification},
  author  = {Rong, Yu and Huang, Wenbing and Xu, Tingyang and Huang, Junzhou},
  journal = {arXiv preprint arXiv:1907.10903},
  year    = {2019}
}

@inproceedings{sun2021mocl,
  title     = {{MoCL}: Data-Driven Molecular Fingerprint via Knowledge-Aware Contrastive Learning from Molecular Graph},
  author    = {Sun, Mengying and Xing, Jing and Wang, Huijun and Chen, Bin and Zhou, Jiayu},
  booktitle = {Proceedings of the 27th ACM SIGKDD Conference on Knowledge Discovery \& Data Mining},
  pages     = {3585--3594},
  year      = {2021}
}

@article{ju2023unsupervised,
  title     = {Unsupervised Graph-Level Representation Learning with Hierarchical Contrasts},
  author    = {Ju, Wei and Gu, Yiyang and Luo, Xiao and Wang, Yifan and Yuan, Haochen and Zhong, Huasong and Zhang, Ming},
  journal   = {Neural Networks},
  volume    = {158},
  pages     = {359--368},
  year      = {2023},
  publisher = {Elsevier}
}

@inproceedings{ren2021label,
  title     = {Label Contrastive Coding Based Graph Neural Network for Graph Classification},
  author    = {Ren, Yuxiang and Bai, Jiyang and Zhang, Jiawei},
  booktitle = {Database Systems for Advanced Applications},
  pages     = {123--140},
  year      = {2021},
  publisher = {Springer}
}

@inproceedings{park2020unsupervised,
  title     = {Unsupervised Attributed Multiplex Network Embedding},
  author    = {Park, Chanyoung and Kim, Donghyun and Han, Jiawei and Yu, Hwanjo},
  booktitle = {Proceedings of the AAAI Conference on Artificial Intelligence},
  volume    = {34},
  pages     = {5371--5378},
  year      = {2020}
}

@article{hjelm2018learning,
  title   = {Learning Deep Representations by Mutual Information Estimation and Maximization},
  author  = {Hjelm, R. Devon and Fedorov, Alex and Lavoie-Marchildon, Samuel and Grewal, Karan and Bachman, Philip and Trischler, Adam and Bengio, Yoshua},
  journal = {arXiv preprint arXiv:1808.06670},
  year    = {2018}
}

@inproceedings{xia2022progcl,
  title     = {{ProGCL}: Rethinking Hard Negative Mining in Graph Contrastive Learning},
  author    = {Xia, Jun and Wu, Lirong and Wang, Ge and Chen, Jintao and Li, Stan Z.},
  booktitle = {International Conference on Machine Learning},
  pages     = {24332--24346},
  year      = {2022},
  publisher = {PMLR}
}

@article{zhang2022localized,
  title   = {Localized Contrastive Learning on Graphs},
  author  = {Zhang, Hengrui and Wu, Qitian and Wang, Yu and Zhang, Shaofeng and Yan, Junchi and Yu, Philip S.},
  journal = {arXiv preprint arXiv:2212.04604},
  year    = {2022}
}

@book{edgar2008measure,
  title     = {Measure, Topology, and Fractal Geometry},
  author    = {Edgar, Gerald A.},
  volume    = {2},
  year      = {2008},
  publisher = {Springer}
}

@book{mandelbrot1983fractal,
  title     = {The Fractal Geometry of Nature},
  author    = {Mandelbrot, Benoit B.},
  year      = {1983},
  publisher = {W. H. Freeman}
}

@article{mandelbrot1989fractal,
  title     = {Fractal Geometry: What Is It, and What Does It Do?},
  author    = {Mandelbrot, Benoit B.},
  journal   = {Proceedings of the Royal Society of London. A. Mathematical and Physical Sciences},
  volume    = {423},
  number    = {1864},
  pages     = {3--16},
  year      = {1989},
  publisher = {The Royal Society}
}

@article{neroli2024fractal,
  title     = {Fractal Dimensions for Iterated Graph Systems},
  author    = {Neroli, Ziyu},
  journal   = {Proceedings of the Royal Society A},
  volume    = {480},
  number    = {2300},
  pages     = {20240406},
  year      = {2024},
  publisher = {The Royal Society}
}

@article{zakar2023towards,
  title     = {Towards a Better Understanding of the Characteristics of Fractal Networks},
  author    = {Zakar-Poly{\'a}k, Enik{\H{o}} and Nagy, Marcell and Molontay, Roland},
  journal   = {Applied Network Science},
  volume    = {8},
  number    = {17},
  pages     = {1--34},
  year      = {2023},
  publisher = {SpringerOpen}
}

@article{ma2020understanding,
  title   = {Understanding Chinese Urban Form: The Universal Fractal Pattern of Street Networks over 298 Cities},
  author  = {Ma, Ding and Guo, Renzhong and Zheng, Ye and Zhao, Zhigang and He, Fangning and Zhu, Wei},
  journal = {ISPRS International Journal of Geo-Information},
  volume  = {9},
  number  = {4},
  pages   = {192},
  year    = {2020},
  doi     = {10.3390/ijgi9040192}
}

@article{laurienti2011universal,
  title   = {Universal Fractal Scaling of Self-Organized Networks},
  author  = {Laurienti, Paul J. and Joyce, Karen E. and Telesford, Qawi K. and Burdette, Jonathan H. and Hayasaka, Satoru},
  journal = {Physica A: Statistical Mechanics and its Applications},
  volume  = {390},
  number  = {20},
  pages   = {3608--3613},
  year    = {2011},
  doi     = {10.1016/j.physa.2011.05.011}
}

@article{song2005self,
  title     = {Self-Similarity of Complex Networks},
  author    = {Song, Chaoming and Havlin, Shlomo and Makse, Hern{\'a}n A.},
  journal   = {Nature},
  volume    = {433},
  number    = {7024},
  pages     = {392--395},
  year      = {2005},
  publisher = {Nature Publishing Group}
}

@inproceedings{luo2025can,
  title     = {Can Classic GNNs Be Strong Baselines for Graph-Level Tasks? Simple Architectures Meet Excellence},
  author    = {Luo, Yuankai and Shi, Lei and Wu, Xiao-Ming},
  booktitle = {International Conference on Machine Learning},
  year      = {2025}
}

@article{wang2024graph,
  title   = {Graph-Mamba: Towards Long-Range Graph Sequence Modeling with Selective State Spaces},
  author  = {Wang, Chloe and Tsepa, Oleksii and Ma, Jun and Wang, Bo},
  journal = {arXiv preprint arXiv:2402.00789},
  year    = {2024}
}

@article{morris2020tudataset,
  title   = {{TUDataset}: A Collection of Benchmark Datasets for Learning with Graphs},
  author  = {Morris, Christopher and Kriege, Nils M. and Bause, Franka and Kersting, Kristian and Mutzel, Petra and Neumann, Marion},
  journal = {arXiv preprint arXiv:2007.08663},
  year    = {2020}
}

@article{kipf2016variational,
  title   = {Variational Graph Auto-Encoders},
  author  = {Kipf, Thomas N. and Welling, Max},
  journal = {arXiv preprint arXiv:1611.07308},
  year    = {2016}
}

@article{sun2019infograph,
  title   = {{InfoGraph}: Unsupervised and Semi-Supervised Graph-Level Representation Learning via Mutual Information Maximization},
  author  = {Sun, Fan-Yun and Hoffmann, Jordan and Verma, Vikas and Tang, Jian},
  journal = {arXiv preprint arXiv:1908.01000},
  year    = {2019}
}

@inproceedings{li2022let,
  title     = {Let Invariant Rationale Discovery Inspire Graph Contrastive Learning},
  author    = {Li, Sihang and Wang, Xiang and Zhang, An and Wu, Yingxin and He, Xiangnan and Chua, Tat-Seng},
  booktitle = {International Conference on Machine Learning},
  pages     = {13052--13065},
  year      = {2022},
  publisher = {PMLR}
}

@inproceedings{ji2024rethinking,
  title     = {Rethinking Dimensional Rationale in Graph Contrastive Learning from Causal Perspective},
  author    = {Ji, Qirui and Li, Jiangmeng and Hu, Jie and Wang, Rui and Zheng, Changwen and Xu, Fanjiang},
  booktitle = {Proceedings of the AAAI Conference on Artificial Intelligence},
  volume    = {38},
  pages     = {12810--12820},
  year      = {2024}
}

@inproceedings{li2024gradgcl,
  title     = {{GradGCL}: Gradient Graph Contrastive Learning},
  author    = {Li, Ran and Di, Shimin and Chen, Lei and Zhou, Xiaofang},
  booktitle = {2024 IEEE 40th International Conference on Data Engineering},
  pages     = {1171--1184},
  year      = {2024},
  publisher = {IEEE}
}

@article{zhai2025heterogeneous,
  title     = {Heterogeneous Graph Neural Networks with Post-Hoc Explanations for Multi-Modal and Explainable Land Use Inference},
  author    = {Zhai, Xuehao and Jiang, Junqi and Dejl, Adam and Rago, Antonio and Guo, Fangce and Toni, Francesca and Sivakumar, Aruna},
  journal   = {Information Fusion},
  volume    = {120},
  pages     = {103057},
  year      = {2025},
  publisher = {Elsevier}
}

@article{zhao2024exploring,
  title     = {Exploring the Impact of Trip Patterns on Spatially Aggregated Crashes Using Floating Vehicle Trajectory Data and Graph Convolutional Networks},
  author    = {Zhao, Jiahui and Liu, Pan and Li, Zhibin},
  journal   = {Accident Analysis \& Prevention},
  volume    = {194},
  pages     = {107340},
  year      = {2024},
  publisher = {Elsevier}
}

@article{watts1998collective,
  title     = {Collective Dynamics of {`Small-World'} Networks},
  author    = {Watts, Duncan J. and Strogatz, Steven H.},
  journal   = {Nature},
  volume    = {393},
  number    = {6684},
  pages     = {440--442},
  year      = {1998},
  publisher = {Nature Publishing Group}
}

@article{narayanan2017graph2vec,
  title={graph2vec: Learning distributed representations of graphs},
  author={Narayanan, Annamalai and Chandramohan, Mahinthan and Venkatesan, Rajasekar and Chen, Lihui and Liu, Yang and Jaiswal, Shantanu},
  journal={arXiv preprint arXiv:1707.05005},
  year={2017}
}

@article{barabasi1999emergence,
  title     = {Emergence of Scaling in Random Networks},
  author    = {Barab{\'a}si, Albert-L{\'a}szl{\'o} and Albert, R{\'e}ka},
  journal   = {Science},
  volume    = {286},
  number    = {5439},
  pages     = {509--512},
  year      = {1999},
  publisher = {American Association for the Advancement of Science}
}

@article{griffiths1982spin,
  title     = {Spin Systems on Hierarchical Lattices. Introduction and Thermodynamic Limit},
  author    = {Griffiths, Robert B. and Kaufman, Miron},
  journal   = {Physical Review B},
  volume    = {26},
  number    = {9},
  pages     = {5022},
  year      = {1982},
  publisher = {APS}
}

@article{li2024scale,
  title   = {On the Scale-Freeness of Random Colored Substitution Networks},
  author  = {Li, Nero Ziyu and Britz, Thomas},
  journal = {Proceedings of the American Mathematical Society},
  volume  = {152},
  number  = {4},
  pages   = {1377--1389},
  year    = {2024}
}

@inproceedings{wei2023boosting,
  title     = {Boosting Graph Contrastive Learning via Graph Contrastive Saliency},
  author    = {Wei, Chunyu and Wang, Yu and Bai, Bing and Ni, Kai and Brady, David and Fang, Lu},
  booktitle = {International Conference on Machine Learning},
  pages     = {36839--36855},
  year      = {2023},
  publisher = {PMLR}
}

@article{jin2021multi,
  title   = {Multi-Scale Contrastive Siamese Networks for Self-Supervised Graph Representation Learning},
  author  = {Jin, Ming and Zheng, Yizhen and Li, Yuan-Fang and Gong, Chen and Zhou, Chuan and Pan, Shirui},
  journal = {arXiv preprint arXiv:2105.05682},
  year    = {2021}
}

@inproceedings{velickovic2019deep,
  title = {Deep Graph Infomax},
  author = {Veli{\v{c}}kovi{\'c}, Petar and Fedus, William and Hamilton, William L. and Li{\`o}, Pietro and Bengio, Yoshua and Hjelm, R. Devon},
  booktitle = {International Conference on Learning Representations},
  year = {2019},
  url = {https://openreview.net/forum?id=rklz9iAcKQ}
}

@inproceedings{sun2020infograph,
  title = {{InfoGraph}: Unsupervised and Semi-supervised Graph-Level Representation Learning via Mutual Information Maximization},
  author = {Sun, Fan-Yun and Hoffmann, Jordan and Verma, Vikas and Tang, Jian},
  booktitle = {International Conference on Learning Representations},
  year = {2020},
  url = {https://openreview.net/forum?id=r1lfF2NYvH}
}

@inproceedings{qiu2020gcc,
  title = {{GCC}: Graph Contrastive Coding for Graph Neural Network Pre-Training},
  author = {Qiu, Jiezhong and Chen, Qibin and Dong, Yuxiao and Zhang, Jing and Yang, Hongxia and Ding, Ming and Wang, Kuansan and Tang, Jie},
  booktitle = {Proceedings of the 26th ACM SIGKDD International Conference on Knowledge Discovery and Data Mining},
  year = {2020},
  pages = {1150--1160},
  doi = {10.1145/3394486.3403168},
  url = {https://doi.org/10.1145/3394486.3403168}
}

@inproceedings{you2020graph,
  title = {Graph Contrastive Learning with Augmentations},
  author = {You, Yuning and Chen, Tianlong and Sui, Yongduo and Chen, Ting and Wang, Zhangyang and Shen, Yang},
  booktitle = {Advances in Neural Information Processing Systems},
  volume = {33},
  year = {2020},
  url = {https://proceedings.neurips.cc/paper/2020/hash/3fe230348e9a12c13120749e3f9fa4cd-Abstract.html}
}

@inproceedings{suresh2021adversarial,
  title = {Adversarial Graph Augmentation to Improve Graph Contrastive Learning},
  author = {Suresh, Susheel and Li, Pan and Hao, Cong and Neville, Jennifer},
  booktitle = {Advances in Neural Information Processing Systems},
  volume = {34},
  year = {2021},
  url = {https://openreview.net/forum?id=ioyq7NsR1KJ}
}

@article{zhang2024personalized,
  title = {Graph Contrastive Learning with Personalized Augmentation},
  author = {Zhang, Xin and Tan, Qiaoyu and Huang, Xiao and Li, Bo},
  journal = {IEEE Transactions on Knowledge and Data Engineering},
  volume = {36},
  number = {11},
  pages = {6305--6316},
  year = {2024},
  doi = {10.1109/TKDE.2024.3388728},
  url = {https://doi.org/10.1109/TKDE.2024.3388728}
}

@inproceedings{zhuo2024unified,
  title = {Unified Graph Augmentations for Generalized Contrastive Learning on Graphs},
  author = {Zhuo, Jiaming and Lu, Yintong and Ning, Hui and Fu, Kun and Niu, Bingxin and He, Dongxiao and Wang, Chuan and Guo, Yuanfang and Wang, Zhen and Cao, Xiaochun and Yang, Liang},
  booktitle = {Advances in Neural Information Processing Systems},
  volume = {37},
  year = {2024},
  doi = {10.52202/079017-1183},
  url = {https://proceedings.neurips.cc/paper_files/paper/2024/hash/41efc12982eca6f8bb5e48dc3a84b843-Abstract-Conference.html}
}

@inproceedings{he2024representation,
  title = {Exploitation of a Latent Mechanism in Graph Contrastive Learning: Representation Scattering},
  author = {He, Dongxiao and Shan, Lianze and Zhao, Jitao and Zhang, Hengrui and Wang, Zhen and Zhang, Weixiong},
  booktitle = {Advances in Neural Information Processing Systems},
  volume = {37},
  year = {2024},
  url = {https://proceedings.neurips.cc/paper_files/paper/2024/hash/d0ffb35aaa7faa894afe5060c694d674-Abstract-Conference.html}
}

@inproceedings{liu2024reinforcement,
  title = {Graph Contrastive Learning with Reinforcement Augmentation},
  author = {Liu, Ziyang and Wang, Chaokun and Wu, Cheng},
  booktitle = {Proceedings of the Thirty-Third International Joint Conference on Artificial Intelligence},
  pages = {2225--2233},
  year = {2024},
  doi = {10.24963/ijcai.2024/246},
  url = {https://www.ijcai.org/proceedings/2024/246}
}

@misc{choi2025fractal,
  title = {Fractal-Inspired Message Passing Neural Networks with Fractal Nodes},
  author = {Choi, Jeongwhan and Park, Seungjun and Park, Sumin and Cho, Sung-Bae and Park, Noseong},
  year = {2025},
  note = {OpenReview preprint},
  url = {https://openreview.net/forum?id=zPoW8CajCN}
}

@misc{choi2026are,
  title = {Are Graph Transformers Necessary? Efficient Long-Range Message Passing with Fractal Nodes in MPNNs},
  author = {Choi, Jeongwhan and Park, Seungjun and Park, Sumin and Cho, Sung-Bae and Park, Noseong},
  year = {2025},
  eprint = {2511.13010},
  archivePrefix = {arXiv},
  primaryClass = {cs.LG},
  url = {https://arxiv.org/abs/2511.13010}
}

@inproceedings{cao2023gst,
  title = {Learning Large Graph Property Prediction via Graph Segment Training},
  author = {Cao, Kaidi and Phothilimthana, Mangpo and Abu-El-Haija, Sami and Zelle, Dustin and Zhou, Yanqi and Mendis, Charith and Leskovec, Jure and Perozzi, Bryan},
  booktitle = {Advances in Neural Information Processing Systems},
  volume = {36},
  year = {2023},
  url = {https://proceedings.neurips.cc/paper_files/paper/2023/hash/48f8143cebe113f4596e1781771578cd-Abstract-Conference.html}
}

@misc{lo2022android,
  title = {Graph Neural Network-based Android Malware Classification with Jumping Knowledge},
  author = {Lo, Wai Weng and Layeghy, Siamak and Sarhan, Mohanad and Gallagher, Marcus and Portmann, Marius},
  year = {2022},
  eprint = {2201.07537},
  archivePrefix = {arXiv},
  primaryClass = {cs.CR},
  url = {https://arxiv.org/abs/2201.07537}
}

@article{azizpour2025model,
  title={Model-Driven Graph Contrastive Learning},
  author={Azizpour, Ali and Zilberstein, Nicolas and Segarra, Santiago},
  journal={arXiv preprint arXiv:2506.06212},
  year={2025}
}

\section*{Appendix}

\appendix

\section{Theoretical Details and Proofs}\label{appendix: math}





This appendix gives the formal version of the box-counting, renormalisation, and stochastic-surrogate statements used in Section~\ref{subsection:background-of-fractal-learning}--Section~\ref{subsection:implementation}.  We distinguish the limiting dimension $\dimb(G^\infty)$ of a graph sequence from the finite-scale estimator $\tdimb(G)$ used on a finite graph.

\begin{definition}[Box-counting dimension]\label{definition: box_dimension}
Let $G=(V,E)$ be a connected graph with shortest-path metric $d_G$.  For an integer $L\ge1$, an $L$-box partition is a family of nonempty vertex sets $\{U_i\}_{i\in\mathcal I}$ satisfying
\[
V=\bigsqcup_{i\in\mathcal I}U_i,
\qquad
\diam(U_i):=\sup_{u,v\in U_i}d_G(u,v)\le L .
\]
Let $N_L(G)$ be the minimum number of boxes in such a partition.
For a vertex-increasing sequence $G^\infty=(G^n)_{n\ge1}$ with $\diam(G^n)\to\infty$, define
\[
\dimb(G^\infty)
:=
\lim_{\substack{n\to\infty\\ L_n/\diam(G^n)\to0}}
\frac{\log N_{L_n}(G^n)}{-\log\bigl(L_n/\diam(G^n)\bigr)},
\]
provided the limit exists and is independent of the admissible integer scale sequence $(L_n)_{n\ge1}$.  For a finite graph $G$, the estimator $\tdimb(G)$ is the least-squares slope in
\[
\log N_L(G)
=
 b_G+
\tdimb(G)\left[-\log\left(\frac{L}{\diam(G)}\right)\right]+
\varepsilon_L,
\qquad
L\in\{1,\ldots,\lfloor\diam(G)/2\rfloor\}.
\]
The associated coefficient of determination is denoted by $R^2(G)$.
\end{definition}

\begin{algorithm}[ht]
\caption{Algorithm of computing box dimension}
\label{appendix: algorithm_dimension}
\begin{algorithmic}[1]
\Require Graph $\mathcal{G}$ with node set $\mathcal{V}$.
\Ensure Fractality metric $R^2$ and finite-scale box dimension $\tdimb(\mathcal{G})$.
\If{$\diam(\mathcal{G}) \le 9$} \Comment{too small for fractal analysis}
  \State $R^2 \gets 0$.
  \State $\tdimb(\mathcal{G}) \gets 0$.
\Else
  \State $L_{\max} \gets \lfloor \diam(\mathcal{G})/2 \rfloor$.
  \State $\textit{Array} \gets \varnothing$.
  \For{$L \gets 1$ to $L_{\max}$}
    \State $r \gets \lfloor L/2 \rfloor$.
    \State $\mathcal{V}_{\mathrm{remain}} \gets \mathcal{V}$.
    \State $N_B(L) \gets 0$.
    \If{$L$ is even}
      \While{$\mathcal{V}_{\mathrm{remain}} \ne \varnothing$}
        \State $v \gets \arg\max_{v \in \mathcal{V}_{\mathrm{remain}}} |B(v,r)|$.
        \State $\mathcal{V}_{\mathrm{remain}} \gets \mathcal{V}_{\mathrm{remain}} \setminus B(v,r)$.
        \State $N_B(L) \gets N_B(L) + 1$.
      \EndWhile
    \Else
      \While{$\mathcal{V}_{\mathrm{remain}} \ne \varnothing$}
        \If{$\exists v,w \in \mathcal{V}_{\mathrm{remain}} \text{ with } d_{\mathcal{G}}(v,w)=1$}
          \State $(v,w) \gets \arg\max_{d_{\mathcal{G}}(v,w)=1} |B(v,r)\cup B(w,r)|$.
          \State $\mathcal{V}_{\mathrm{remain}} \gets \mathcal{V}_{\mathrm{remain}} \setminus \big(B(v,r)\cup B(w,r)\big)$.
          \State $N_B(L) \gets N_B(L) + 1$.
        \Else
          \State $v \gets \arg\max_{v \in \mathcal{V}_{\mathrm{remain}}} |B(v,r)|$.
          \State $\mathcal{V}_{\mathrm{remain}} \gets \mathcal{V}_{\mathrm{remain}} \setminus B(v,r)$.
          \State $N_B(L) \gets N_B(L) + 1$.
        \EndIf
      \EndWhile
    \EndIf
    \State $\textit{Array} \gets \textit{Array} \cup \{(-\log(L/\diam(\mathcal{G})),\ \log N_B(L))\}$.
  \EndFor
  \State Fit $y = m x + b$ to $\textit{Array}$ by least squares and compute $R^2$.
  \State $\tdimb(\mathcal{G}) \gets m$.
\EndIf
\State \Return $R^2,\ \tdimb(\mathcal{G})$.
\end{algorithmic}
\end{algorithm}

\begin{algorithm}[ht]
\caption{Algorithm of random-centre renormalisation}
\label{appendix: algorithm_renormalisation}
\begin{algorithmic}[1]
\State \textbf{Input:} Graph $G$ with node set $\mathcal{V}$ and adjacency matrix $A$, radius $r$.
\State \textbf{Output:} Renormalised graph $\mathscr{R}(G)$.
\State $\mathcal{V}_{\mathrm{remain}}\gets\mathcal{V}$, $\mathcal{V}_{\mathrm{super}}\gets\varnothing$.
\State $A_r\gets\sum_{i=1}^r A^i$.
\While{$\mathcal{V}_{\mathrm{remain}}\neq\varnothing$}
  \State Pick $u$ uniformly from $\mathcal{V}_{\mathrm{remain}}$.
  \State $U\gets\{i\in\mathcal{V}_{\mathrm{remain}}:A_r[u][i]>0\}\cup\{u\}$.
  \State Add $U$ to $\mathcal{V}_{\mathrm{super}}$.
  \State $\mathcal{V}_{\mathrm{remain}}\gets\mathcal{V}_{\mathrm{remain}}\setminus U$.
\EndWhile
\State Build an assignment matrix $S$ with $S_{kj}=1$ iff vertex $j$ belongs to the $k$-th supervertex.
\State $A_{\mathrm{super}}\gets S A S^\top$.
\State Binarise $A_{\mathrm{super}}$ and remove diagonal entries.
\State \Return $\mathscr{R}(G)$ defined by $A_{\mathrm{super}}$.
\end{algorithmic}
\end{algorithm}

\begin{theorem}
\label{appendix: estimator_consistency}
Let $(G^n,d_{G^n}/\diam(G^n))$ converge to a compact Gromov--Hausdorff scaling limit $X$ whose box dimension exists.  If the discrete box-counting curves converge uniformly to the limiting box-counting curve on every compact scale interval used by the regression, then
\[
\tdimb(G^n)\longrightarrow \dimb(X)=\dimb(G^\infty).
\]
\end{theorem}
\begin{proof}
Let $x_L^n=-\log(L/\diam(G^n))$ and $y_L^n=\log N_L(G^n)$.  The finite estimator $\tdimb(G^n)$ is the minimiser of the least-squares functional for the empirical pairs $(x_L^n,y_L^n)$.  Uniform convergence of the discrete box-counting curves to the limiting box-counting curve implies uniform convergence of these least-squares functionals on bounded slope intervals.  Since the limiting curve has slope $\dimb(X)$ and the limiting least-squares minimiser is unique under non-degenerate scale spread, the argmin convergence theorem yields $\tdimb(G^n)\to\dimb(X)$.  The equality $\dimb(X)=\dimb(G^\infty)$ follows from the definition of the scaling limit and the admissible scale normalisation.
\end{proof}

\begin{theorem}[Theorem~\ref{theorem: renormalisation_keep_dimension}]
\label{appendix: theorem_renormalisation_keep_dimension}
Let $G^\infty=(G^n)_{n\ge1}$ be a vertex-increasing sequence of finite connected graphs with $\diam(G^n)\to\infty$.  For every $n$, suppose $\mathscr{R}(G^n)$ is obtained by collapsing a disjoint connected box partition of $G^n$ whose box diameters are bounded by the same finite constant $L_0$.  Let $\mathscr{R}(G^\infty)=(\mathscr{R}(G^n))_{n\ge1}$.  If the box dimensions of both graph sequences exist under Definition~\ref{definition: box_dimension}, then
\[
\dimb\bigl(\mathscr{R}(G^\infty)\bigr)=\dimb(G^\infty).
\]
\end{theorem}

\begin{proof}
For each $n$, let $\pi_n:V(G^n)\to V(\mathscr{R}(G^n))$ be the collapse map sending a vertex to the supervertex corresponding to its assigned box.  Because the boxes are disjoint, $\pi_n$ is well defined.  Every path in $G^n$ projects to a path in $\mathscr{R}(G^n)$ after deleting repeated consecutive supervertices, so
\[
d_{\mathscr{R}(G^n)}(\pi_n(x),\pi_n(y))\le d_{G^n}(x,y).
\]
Conversely, a quotient path of length $m$ can be lifted to $G^n$ by crossing one original edge between adjacent boxes and travelling inside boxes of diameter at most $L_0$, giving
\[
d_{G^n}(x,y)\le (L_0+1)d_{\mathscr{R}(G^n)}(\pi_n(x),\pi_n(y))+L_0 .
\]
Thus the two metrics are quasi-isometric with constants independent of $n$, and their diameters are comparable up to multiplicative and additive constants independent of $n$.

An $L$-box partition of $G^n$ projects to an $L$-box cover of $\mathscr{R}(G^n)$, hence $N_L(\mathscr{R}(G^n))\le N_L(G^n)$.  Conversely, the inverse image of an $L$-box partition of $\mathscr{R}(G^n)$ is a partition of $G^n$ by boxes of diameter at most $(L_0+1)L+L_0$, hence
\[
N_{(L_0+1)L+L_0}(G^n)
\le
N_L(\mathscr{R}(G^n))
\le
N_L(G^n).
\]
For any admissible scale sequence, the additive and multiplicative constants above contribute only bounded shifts to the logarithmic normalisers.  Passing to the limit in Definition~\ref{definition: box_dimension} gives $\dimb(\mathscr{R}(G^\infty))=\dimb(G^\infty)$.
\end{proof}

\paragraph{Effect of the fractal term.}
We next record how the box-counting term changes the repulsive force assigned
to negative pairs.  For an anchor graph \(G_i\), the FractalGCL loss is
\[
\ell_i
=
-\log
\frac{
\exp\left(
\operatorname{sim}
\left(
\mathbf z_i,
\mathbf z_i^{(\mathcal A)}
\right)/\tau
-
\alpha
\left|
\widetilde{\dim}_{\mathrm B}(G_i)
-
\widetilde{\dim}_{\mathrm B}\left(\mathscr R(G_i)\right)
\right|
\right)
}{
\sum_{\substack{j=1\\j\neq i}}^{N}
\exp\left(
\operatorname{sim}
\left(
\mathbf z_i,
\mathbf z_j^{(\mathcal A)}
\right)/\tau
+
\alpha
\left|
\widetilde{\dim}_{\mathrm B}(G_i)
-
\widetilde{\dim}_{\mathrm B}\left(\mathscr R(G_j)\right)
\right|
\right)
}.
\]

\begin{theorem}
\label{appendix:theorem_fractal_negative_repulsion}
For every \(j\neq i\),
\[
\frac{
\partial \ell_i
}{
\partial
\operatorname{sim}\left(
\mathbf z_i,
\mathbf z_j^{(\mathcal A)}
\right)
}
=
\frac{1}{\tau}
\frac{
\exp\left(
\operatorname{sim}\left(
\mathbf z_i,
\mathbf z_j^{(\mathcal A)}
\right)/\tau
+
\alpha
\left|
\widetilde{\dim}_{\mathrm B}(G_i)
-
\widetilde{\dim}_{\mathrm B}\left(\mathscr R(G_j)\right)
\right|
\right)
}{
\sum_{\substack{k=1\\k\neq i}}^{N}
\exp\left(
\operatorname{sim}\left(
\mathbf z_i,
\mathbf z_k^{(\mathcal A)}
\right)/\tau
+
\alpha
\left|
\widetilde{\dim}_{\mathrm B}(G_i)
-
\widetilde{\dim}_{\mathrm B}\left(\mathscr R(G_k)\right)
\right|
\right)
}.
\]
Moreover, for \(j\neq i\) and \(k\neq i\),
\begin{align*}
\frac{
\partial \ell_i /
\partial
\operatorname{sim}\left(
\mathbf z_i,
\mathbf z_j^{(\mathcal A)}
\right)
}{
\partial \ell_i /
\partial
\operatorname{sim}\left(
\mathbf z_i,
\mathbf z_k^{(\mathcal A)}
\right)
}
&=
\exp\left(
\begin{aligned}
&
\frac{
\operatorname{sim}\left(
\mathbf z_i,
\mathbf z_j^{(\mathcal A)}
\right)
-
\operatorname{sim}\left(
\mathbf z_i,
\mathbf z_k^{(\mathcal A)}
\right)
}{\tau}
\\
&+
\alpha
\left[
\left|
\widetilde{\dim}_{\mathrm B}(G_i)
-
\widetilde{\dim}_{\mathrm B}\left(\mathscr R(G_j)\right)
\right|
-
\left|
\widetilde{\dim}_{\mathrm B}(G_i)
-
\widetilde{\dim}_{\mathrm B}\left(\mathscr R(G_k)\right)
\right|
\right]
\end{aligned}
\right).
\end{align*}
\end{theorem}

\begin{proof}
Expanding the logarithm gives
\[
\ell_i
=
-
\frac{
\operatorname{sim}
\left(
\mathbf z_i,
\mathbf z_i^{(\mathcal A)}
\right)
}{\tau}
+
\alpha
\left|
\widetilde{\dim}_{\mathrm B}(G_i)
-
\widetilde{\dim}_{\mathrm B}\left(\mathscr R(G_i)\right)
\right|
\]
\[
+
\log
\sum_{\substack{k=1\\k\neq i}}^{N}
\exp\left(
\operatorname{sim}
\left(
\mathbf z_i,
\mathbf z_k^{(\mathcal A)}
\right)/\tau
+
\alpha
\left|
\widetilde{\dim}_{\mathrm B}(G_i)
-
\widetilde{\dim}_{\mathrm B}\left(\mathscr R(G_k)\right)
\right|
\right).
\]
For \(j\neq i\), the term
\[
\operatorname{sim}\left(
\mathbf z_i,
\mathbf z_j^{(\mathcal A)}
\right)
\]
appears only inside the log-sum-exp term. Therefore,
\[
\frac{
\partial \ell_i
}{
\partial
\operatorname{sim}\left(
\mathbf z_i,
\mathbf z_j^{(\mathcal A)}
\right)
}
=
\frac{1}{\tau}
\frac{
\exp\left(
\operatorname{sim}\left(
\mathbf z_i,
\mathbf z_j^{(\mathcal A)}
\right)/\tau
+
\alpha
\left|
\widetilde{\dim}_{\mathrm B}(G_i)
-
\widetilde{\dim}_{\mathrm B}\left(\mathscr R(G_j)\right)
\right|
\right)
}{
\sum_{\substack{k=1\\k\neq i}}^{N}
\exp\left(
\operatorname{sim}\left(
\mathbf z_i,
\mathbf z_k^{(\mathcal A)}
\right)/\tau
+
\alpha
\left|
\widetilde{\dim}_{\mathrm B}(G_i)
-
\widetilde{\dim}_{\mathrm B}\left(\mathscr R(G_k)\right)
\right|
\right)
}.
\]
Dividing the derivative for index \(j\) by the derivative for index \(k\)
cancels the common denominator and the factor \(1/\tau\), which gives the ratio
formula.
\end{proof}

The theorem gives the exact gradient-level role of the fractal term. Different
graphs remain negative samples. The box-counting term does not directly turn
box-counting-similar graphs into positives; it changes the strength of negative
repulsion.

\begin{corollary}
\label{corollary:larger_discrepancy_stronger_repulsion}
If
\[
\operatorname{sim}\left(
\mathbf z_i,
\mathbf z_j^{(\mathcal A)}
\right)
=
\operatorname{sim}\left(
\mathbf z_i,
\mathbf z_k^{(\mathcal A)}
\right)
\]
and
\[
\left|
\widetilde{\dim}_{\mathrm B}(G_i)
-
\widetilde{\dim}_{\mathrm B}\left(\mathscr R(G_j)\right)
\right|
>
\left|
\widetilde{\dim}_{\mathrm B}(G_i)
-
\widetilde{\dim}_{\mathrm B}\left(\mathscr R(G_k)\right)
\right|,
\]
then
\[
\frac{
\partial \ell_i
}{
\partial
\operatorname{sim}\left(
\mathbf z_i,
\mathbf z_j^{(\mathcal A)}
\right)
}
>
\frac{
\partial \ell_i
}{
\partial
\operatorname{sim}\left(
\mathbf z_i,
\mathbf z_k^{(\mathcal A)}
\right)
}.
\]
\end{corollary}

\begin{proof}
Under the equal-similarity condition, Theorem~\ref{appendix:theorem_fractal_negative_repulsion}
gives
\[
\frac{
\partial \ell_i /
\partial
\operatorname{sim}\left(
\mathbf z_i,
\mathbf z_j^{(\mathcal A)}
\right)
}{
\partial \ell_i /
\partial
\operatorname{sim}\left(
\mathbf z_i,
\mathbf z_k^{(\mathcal A)}
\right)
}
=
\exp\left(
\alpha
\left[
\left|
\widetilde{\dim}_{\mathrm B}(G_i)
-
\widetilde{\dim}_{\mathrm B}\left(\mathscr R(G_j)\right)
\right|
-
\left|
\widetilde{\dim}_{\mathrm B}(G_i)
-
\widetilde{\dim}_{\mathrm B}\left(\mathscr R(G_k)\right)
\right|
\right]
\right).
\]
The exponent is positive by assumption and \(\alpha>0\). Hence the ratio is
larger than \(1\).
\end{proof}

\begin{corollary}
\label{corollary:dimension_similar_not_positive}
For every \(j\neq i\),
\[
\frac{
\partial \ell_i
}{
\partial
\operatorname{sim}\left(
\mathbf z_i,
\mathbf z_j^{(\mathcal A)}
\right)
}
>
0.
\]
If
\[
\operatorname{sim}\left(
\mathbf z_i,
\mathbf z_j^{(\mathcal A)}
\right)
=
\operatorname{sim}\left(
\mathbf z_i,
\mathbf z_k^{(\mathcal A)}
\right)
\]
and
\[
\left|
\widetilde{\dim}_{\mathrm B}(G_i)
-
\widetilde{\dim}_{\mathrm B}\left(\mathscr R(G_j)\right)
\right|
<
\left|
\widetilde{\dim}_{\mathrm B}(G_i)
-
\widetilde{\dim}_{\mathrm B}\left(\mathscr R(G_k)\right)
\right|,
\]
then
\[
\frac{
\partial \ell_i
}{
\partial
\operatorname{sim}\left(
\mathbf z_i,
\mathbf z_j^{(\mathcal A)}
\right)
}
<
\frac{
\partial \ell_i
}{
\partial
\operatorname{sim}\left(
\mathbf z_i,
\mathbf z_k^{(\mathcal A)}
\right)
}.
\]
\end{corollary}

\begin{proof}
The first claim follows from Theorem~\ref{appendix:theorem_fractal_negative_repulsion},
because the exponential term, the denominator, and \(\tau\) are all positive.
The second claim follows from the same ratio formula:
\[
\frac{
\partial \ell_i /
\partial
\operatorname{sim}\left(
\mathbf z_i,
\mathbf z_j^{(\mathcal A)}
\right)
}{
\partial \ell_i /
\partial
\operatorname{sim}\left(
\mathbf z_i,
\mathbf z_k^{(\mathcal A)}
\right)
}
=
\exp\left(
\alpha
\left[
\left|
\widetilde{\dim}_{\mathrm B}(G_i)
-
\widetilde{\dim}_{\mathrm B}\left(\mathscr R(G_j)\right)
\right|
-
\left|
\widetilde{\dim}_{\mathrm B}(G_i)
-
\widetilde{\dim}_{\mathrm B}\left(\mathscr R(G_k)\right)
\right|
\right]
\right)
<
1.
\]
\end{proof}

\begin{corollary}
\label{corollary:dimension_dependent_margin}
If \(\ell_i\leq \varepsilon\), then for every \(j\neq i\),
\[
\operatorname{sim}\left(
\mathbf z_i,
\mathbf z_i^{(\mathcal A)}
\right)
-
\operatorname{sim}\left(
\mathbf z_i,
\mathbf z_j^{(\mathcal A)}
\right)
\geq
\tau\alpha
\left[
\left|
\widetilde{\dim}_{\mathrm B}(G_i)
-
\widetilde{\dim}_{\mathrm B}\left(\mathscr R(G_i)\right)
\right|
+
\left|
\widetilde{\dim}_{\mathrm B}(G_i)
-
\widetilde{\dim}_{\mathrm B}\left(\mathscr R(G_j)\right)
\right|
\right]
-
\tau\varepsilon.
\]
\end{corollary}

\begin{proof}
From the expanded loss,
\[
\ell_i
=
-
\frac{
\operatorname{sim}
\left(
\mathbf z_i,
\mathbf z_i^{(\mathcal A)}
\right)
}{\tau}
+
\alpha
\left|
\widetilde{\dim}_{\mathrm B}(G_i)
-
\widetilde{\dim}_{\mathrm B}\left(\mathscr R(G_i)\right)
\right|
\]
\[
+
\log
\sum_{\substack{k=1\\k\neq i}}^{N}
\exp\left(
\operatorname{sim}
\left(
\mathbf z_i,
\mathbf z_k^{(\mathcal A)}
\right)/\tau
+
\alpha
\left|
\widetilde{\dim}_{\mathrm B}(G_i)
-
\widetilde{\dim}_{\mathrm B}\left(\mathscr R(G_k)\right)
\right|
\right).
\]
For any \(j\neq i\),
\[
\log
\sum_{\substack{k=1\\k\neq i}}^{N}
\exp\left(
\operatorname{sim}
\left(
\mathbf z_i,
\mathbf z_k^{(\mathcal A)}
\right)/\tau
+
\alpha
\left|
\widetilde{\dim}_{\mathrm B}(G_i)
-
\widetilde{\dim}_{\mathrm B}\left(\mathscr R(G_k)\right)
\right|
\right)
\]
\[
\geq
\frac{
\operatorname{sim}
\left(
\mathbf z_i,
\mathbf z_j^{(\mathcal A)}
\right)
}{\tau}
+
\alpha
\left|
\widetilde{\dim}_{\mathrm B}(G_i)
-
\widetilde{\dim}_{\mathrm B}\left(\mathscr R(G_j)\right)
\right|.
\]
Therefore,
\[
\ell_i
\geq
\frac{
\operatorname{sim}
\left(
\mathbf z_i,
\mathbf z_j^{(\mathcal A)}
\right)
-
\operatorname{sim}
\left(
\mathbf z_i,
\mathbf z_i^{(\mathcal A)}
\right)
}{\tau}
\]
\[
+
\alpha
\left[
\left|
\widetilde{\dim}_{\mathrm B}(G_i)
-
\widetilde{\dim}_{\mathrm B}\left(\mathscr R(G_i)\right)
\right|
+
\left|
\widetilde{\dim}_{\mathrm B}(G_i)
-
\widetilde{\dim}_{\mathrm B}\left(\mathscr R(G_j)\right)
\right|
\right].
\]
Combining this with \(\ell_i\leq\varepsilon\) and rearranging gives the claim.
\end{proof}

\begin{proposition}
\label{appendix: proposition_fractal_complexity}
Under a cached-neighbourhood implementation of Algorithm~\ref{appendix: algorithm_dimension}, all-scale greedy box covering costs $O(|\mathcal L(G)|\,|V|(|V|+|E|))$.  For sparse graphs this is $O(|V|^3)$ in the worst case, while path-like graphs require $\Omega(|V|^2)$ operations.
\end{proposition}
\begin{proof}
For each scale $L$, cached radius-neighbourhoods can be obtained by truncated breadth-first searches from all vertices, costing $O(|V|(|V|+|E|))$ in the worst case.  The greedy covering at that scale then selects and removes boxes from these cached neighbourhoods, which is no larger in order than the neighbourhood computation.  Summing over $|\mathcal L(G)|$ scales gives $O(|\mathcal L(G)|\,|V|(|V|+|E|))$.  If $|E|=O(|V|)$ and $|\mathcal L(G)|=O(|V|)$, this is $O(|V|^3)$.  For a path graph, at the smallest nontrivial scale the greedy procedure must repeatedly scan a linear number of uncovered vertices while removing only a bounded number per step, giving $\Omega(|V|^2)$ operations.
\end{proof}

\textbf{Auxiliary notation for the stochastic surrogate.}
Let
$
\mathcal L(G)
=
\{1,\ldots,\lfloor \diam(G)/2\rfloor\}
$
be the set of box scales used by the finite-scale estimator, and define
$
x_L(G)
=
-\log(L/\diam(G))
$
for $L\in\mathcal L(G)$.
We also set
\[
S_{xx}\!\left(\diam(G)\right)
=
\sum_{L\in\mathcal L(G)}
\left(
 x_L(G)-\overline{x}_G
\right)^2,
\qquad
\overline{x}_G
=
\frac{1}{|\mathcal L(G)|}
\sum_{L\in\mathcal L(G)}x_L(G).
\]
Let $\widehat\sigma_G^2$ be the residual variance of the log--log box-counting regression used to compute $\tdimb(G)$.

\textbf{Step 1. Finite-diameter slope uncertainty.}

\begin{lemma}
\label{appendix: lemma:_diameter_infinity}
Under the finite-scale box-counting regression model,
\[
\operatorname{SE}(\tdimb(G))=\frac{\widehat\sigma_G}{\sqrt{S_{xx}(\diam(G))}},
\qquad
S_{xx}(\diam(G))\sim\frac{\diam(G)}{2}.
\]
\end{lemma}
\begin{proof}
The OLS slope variance in a simple linear regression with residual variance $\widehat\sigma_G^2$ is $\widehat\sigma_G^2/S_{xx}(\diam(G))$.  It remains to compute the scale spread.  Let $m=\lfloor\diam(G)/2\rfloor$ and $x_L=-\log(L/\diam(G))=\log\diam(G)-\log L$.  Centering removes the constant $\log\diam(G)$, so $S_{xx}=\sum_{L=1}^m(\log L-\overline{\log L})^2$.  Since $\log L=\log m+\log(L/m)+o(1)$ and $L/m$ is asymptotically uniform on $(0,1)$,
\[
\frac1m S_{xx}\to \operatorname{Var}(\log U)=1,
\qquad U\sim\operatorname{Unif}(0,1).
\]
Thus $S_{xx}\sim m\sim \diam(G)/2$, yielding the stated standard error.
\end{proof}

Lemma~\ref{appendix: lemma:_diameter_infinity} shows that the uncertainty of the finite-scale box-counting slope decreases as the graph diameter increases.  This provides a diameter-dependent scale for the renormalisation discrepancy.

\textbf{Step 2. Asymptotic normality of the box-counting slope.}

\begin{lemma}
\label{appendix: lemma_ols_clt}
Under centered Gaussian residuals, or under a Lindeberg-type weak-dependence condition for the log--log box-counting residuals,
\[
\frac{\tdimb(G)-\mathbb E[\tdimb(G)]}{\widehat\sigma_G/\sqrt{S_{xx}(\diam(G))}}
\xrightarrow{\mathscr D}\mathcal N(0,1).
\]
\end{lemma}
\begin{proof}
The OLS slope estimator can be written as a weighted residual sum plus its mean:
\[
\tdimb(G)-\mathbb E[\tdimb(G)]
=
\sum_{L\in\mathcal L(G)}
\frac{x_L-\overline{x}_G}{S_{xx}(\diam(G))}\,\varepsilon_L .
\]
For Gaussian residuals this weighted sum is Gaussian with variance $\widehat\sigma_G^2/S_{xx}(\diam(G))$.  Under the weak-dependence Lindeberg condition, the same standardised weighted sum converges in distribution to $\mathcal N(0,1)$ by the triangular-array central limit theorem.
\end{proof}

\textbf{Step 3. Gaussian surrogate for the renormalisation discrepancy.}
We use the following variance for the stochastic surrogate:
\[
\sigma^2\!\left(\widetilde{\Delta}(G)\right)
=
\lambda_{\Delta}
\frac{\widehat\sigma_G^2}{S_{xx}\!\left(\diam(G)\right)},
\tag{1}
\label{eq:surrogate-variance}
\]
where $\lambda_{\Delta}>0$ is an inflation factor absorbing the uncertainty of the renormalised graph and the dependence between $G$ and $\mathscr R(G)$.
By Lemma~\ref{appendix: lemma:_diameter_infinity},
\[
\sigma^2\!\left(\widetilde{\Delta}(G)\right)
\sim
\frac{2\lambda_{\Delta}\widehat\sigma_G^2}{\diam(G)}.
\tag{2}
\label{eq:diameter-variance}
\]

\begin{proposition}
\label{appendix: proposition_gaussian_surrogate}
Under the finite-scale regression fluctuation model of Lemma~\ref{appendix: lemma_ols_clt}, and with \(\lambda_{\Delta}\) calibrated so that \(\sigma^2(\widetilde{\Delta}(G))\) is asymptotically equivalent to the variance of the centred discrepancy, we have
\[
\frac{\Delta(G)-\mathbb E[\Delta(G)]}{\sigma(\widetilde{\Delta}(G))}
\xrightarrow{\mathscr D}\mathcal N(0,1).
\]
\end{proposition}
\begin{proof}
Write
\[
\Delta(G)=\tdimb(G)-\tdimb(\mathscr R(G)).
\]
By Lemma~\ref{appendix: lemma_ols_clt}, the centred finite-scale estimator of \(G\) is asymptotically normal after its OLS standardisation. The same fluctuation model is assumed for the estimator on \(\mathscr R(G)\). Hence the centred difference of the two estimators is asymptotically normal under joint asymptotic normality. The covariance between the two estimators and the additional uncertainty of the renormalised graph are incorporated into the calibrated inflation factor \(\lambda_\Delta\), which gives the displayed standardisation.

If \((G^n)\) has a Gromov--Hausdorff scaling limit and the finite-scale estimator is consistent as in Theorem~\ref{appendix: estimator_consistency}, then \(\tdimb(G^n)\to\dimb(G^\infty)\). If, in addition, the random renormalisation rule is dimension-consistent and the finite-scale estimators are bounded or uniformly integrable, then this convergence also holds in expectation. Combining this with Theorem~\ref{appendix: theorem_renormalisation_keep_dimension} gives
\[
\mathbb E[\Delta(G^n)]
=
\mathbb E[\tdimb(G^n)-\tdimb(\mathscr R(G^n))]
\to0 .
\]
\end{proof}

The proposition is a conditional approximation rather than a distribution-free theorem for every finite graph. The box-counting residuals of a fixed graph may be correlated across scales, and \(\mathscr R(G)\) is constructed from \(G\), so independence is not assumed as a literal property of the data. The role of \(\lambda_\Delta\) is to absorb the covariance and renormalised-graph uncertainty into a single practical variance scale. Thus the Gaussian surrogate should be interpreted as a diameter-controlled computational approximation justified under the stated fluctuation model.

\section{Related Works}\label{appendix:related}

\textbf{Fractal Geometry for Graphs.}
Fractal geometry interfaces with graph theory most tangibly through the study of complex networks~\citep{watts1998collective,barabasi1999emergence,song2005self}.
Analytical models of genuinely fractal graphs remain comparatively scarce.
Early progress came from physicists who studied percolation and the Ising model on hierarchical lattices and Bethe trees, using real-space renormalisation to obtain non-integer critical exponents and anomalous scaling laws~\citep{griffiths1982spin}.
Recently, the Iterated Graph Systems framework has gained attention for its rigorous yet flexible recursive construction of fractal graphs~\citep{li2024scale,neroli2024fractal}.
However, applications of fractal geometry within graph representation learning are still rare.

\textbf{Graph Contrastive Learning.}
Graph contrastive learning comprises several crucial stages, among which graph data augmentation assumes a pivotal role, yet it is rendered particularly challenging by the intricate non-Euclidean characteristics inherent in graph topologies~\citep{ju2024towards}. Existing graph data augmentation techniques~\citep{velickovic2019deep,you2020graph,you2021graph,qiu2020gcc,li2022let,wei2023boosting,jin2021multi,ji2024rethinking} have achieved notable progress. However, they often fall short in adequately preserving the structural similarity between positive pairs, which arises from the inherent difficulty in precisely leveraging complex topological features.

Graph contrastive learning has become a standard self-supervised paradigm for graph representation learning. Early methods such as Deep Graph Infomax maximize mutual information between local patch representations and global graph summaries, while InfoGraph extends this principle to graph-level representation learning by maximizing the mutual information between graph-level embeddings and multi-scale substructure representations~\citep{velickovic2019deep,sun2020infograph}. GCC further studies contrastive pretraining across graphs by treating sampled subgraphs as instances and learning transferable structural representations~\citep{qiu2020gcc}. GraphCL established a widely used graph-level contrastive protocol: two augmented views of the same graph are treated as a positive pair, while augmented views of different graphs in the same minibatch are treated as negatives~\citep{you2020graph}. Subsequent work improved this paradigm by learning or avoiding augmentations. JOAO formulates augmentation selection as a bi-level optimization problem~\citep{you2021graph}, AD-GCL learns adversarial graph augmentations to reduce redundant information~\citep{suresh2021adversarial}, and SimGRACE removes explicit data augmentation by contrasting representations produced by a clean encoder and a perturbed encoder~\citep{xia2022simgrace}. FractalGCL follows the graph-level contrastive learning principle, but replaces local random perturbations with the renormalisation-based augmented view \(\mathcal A(G)=G\sqcup\mathscr R(G)\) and injects finite-scale box-counting information into the contrastive objective.

\textbf{Recent advances in GCL.}
Recent work has moved beyond manually chosen augmentations and has focused on unified augmentation mechanisms, adaptive view generation, hard negatives, and the internal mechanism of contrastive learning. GOUDA reinterprets graph augmentations from a message-passing perspective and proposes a unified augmentation module that can simulate node, edge, attribute, and subgraph perturbations within a generalized GCL framework~\citep{zhuo2024unified}. Personalized Augmentation observes that different graphs may require different augmentation strategies, rather than applying one fixed strategy to all graphs in a dataset~\citep{zhang2024personalized}. GA2C models graph data augmentation as a Markov decision process and uses a graph advantage actor-critic model to generate progressively evolving augmented views~\citep{liu2024reinforcement}. Another line studies the role of negatives and representation geometry. ProGCL shows that similarity-based hard-negative mining can be unreliable in graphs because many hard negatives are false negatives, and it estimates the probability that a negative is truly negative~\citep{xia2022progcl}. Representation Scattering revisits several GCL frameworks and argues that a common mechanism behind them is scattering representations away from a center while constraining excessive scattering by graph topology~\citep{he2024representation}. FractalGCL is related to these methods because it also changes negative-pair pressure, but its signal is not learned only from embedding similarity or augmentation policies. Instead, the repulsion is modulated by the finite-scale box-counting discrepancy between an anchor graph and the renormalised component of a negative graph, making the loss explicitly sensitive to multi-scale graph geometry.

\textbf{Fractal-inspired graph neural networks and FractalGCL.}
The closest fractal-inspired neural architecture to our work is Fractal-Inspired Message Passing Neural Networks with Fractal Nodes, and its later version on efficient long-range message passing with fractal nodes~\citep{choi2025fractal,choi2026are}. This line of work introduces additional fractal nodes that coexist with original nodes, summarize subgraph-level information, and provide shortcut connections for long-range message passing. Its goal is architectural: improve message passing neural networks by balancing local and global information and alleviating over-squashing. FractalGCL differs in both objective and mechanism. We do not add new node types to the message-passing architecture, and we do not use fractal nodes as shortcut carriers. Instead, we construct a renormalisation-based augmented graph \(\mathcal A(G)=G\sqcup\mathscr R(G)\), estimate finite-scale box-counting dimensions, and use the discrepancy between \(\widetilde{\dim}_{\mathrm B}(G)\) and \(\widetilde{\dim}_{\mathrm B}(\mathscr R(G))\) inside a graph contrastive objective. Thus, Fractal Nodes use fractal structure as an architectural device for long-range propagation, whereas FractalGCL uses renormalisation and box-counting discrepancy as an augmentation-and-loss mechanism for self-supervised graph-level representation learning.

\section{Experimental Methodology and Results of FractalGCL on Urban Districts}

In this section we present the overall experimental framework for evaluating the performance of FractalGCL embeddings on urban districts in three major cities (Chicago, San Francisco and New York).  

\subsection{Setup}

Our pipeline consists of three complementary data modalities extracted for each equal‐area ``catchment'':

\begin{itemize}
  \item \textbf{Road Subgraph Structure:} 
    From the full city road network, we clip each catchment’s local subgraph of nodes and edges, preserving the topological patterns characteristic of that district.  
  \item \textbf{Static Spatial Features:}
    We compute population density and six categories of point‐of‐interest densities (office, sustenance, transportation, retail, leisure and residence), thereby capturing the functional profile of each catchment.  
  \item \textbf{Accident Statistics:}
    Drawing on historical crash data, we aggregate total accident counts and severity level breakdowns to assess safety‐risk characteristics of each catchment.  
\end{itemize}

The high‐level experimental logic proceeds as follows:

\begin{enumerate}
  \item \textbf{Graph Embedding Generation.}
    \begin{itemize}
      \item \emph{FractalGCL Contrastive Training:} We train FractalGCL on the set of catchment subgraphs to produce fixed‐dimensional node and graph embeddings that respect both topology and feature distributions.
      \item \emph{Baseline Encoders:} In parallel, we train several established graph contrastive methods (e.g.\ DGI, InfoGraph, SimGRACE) to serve as performance benchmarks.  
    \end{itemize}

  \item \textbf{Multi‐Task Classification Evaluation.}
    \begin{itemize}
      \item \emph{Accident‐Related Tasks:} We formulate a suite of binary, multi‐class and ordinal classification tasks based on accident counts and severity distributions (e.g.\ high vs.\ low total accidents, severity entropy, risk levels).
    \end{itemize}

  \item \textbf{Performance Comparison and Analysis.}
    \begin{itemize}
      \item For each task, we extract embeddings from each encoder and train a (linear) SVM under repeated stratified cross‐validation.
      \item We compare accuracy and stability metrics across all encoders to quantify the advantages of FractalGCL in integrating topological, functional, and safety information.
    \end{itemize}
\end{enumerate}

\subsection{Hyperparameter Configuration}

In all experiments across Chicago, San Francisco and New York, we used a single, fixed set of hyperparameters for both FractalGCL and the baseline encoders. Specifically, each mini–batch consisted of 16 graph‐level instances, and our GraphSAGE backbone employed two convolutional layers with 64 hidden channels apiece. The final projection head produced 128‐dimensional embeddings for each graph. For contrastive augmentations we applied edge dropping with probability 0.1, and in FractalGCL we injected fractal noise weighted by \(\alpha = 0.4\) after a renormalisation step with radius \(r = 1.0\). All models were trained for 20 epochs using the Adam optimizer with a learning rate of \(10^{-3}\). These settings were held constant to ensure that any observed performance differences arose solely from the encoding method itself rather than hyperparameter variations.

\subsection{Traffic Accident Classification Tasks}

We evaluate each embedding method on six downstream classification tasks based on catchment accident statistics.  Below we list each task name and its precise definition:

\begin{description}
  \item[\texttt{total\_accidents\_high}]  
    \hfill \\
    \emph{Binary classification:} label = 1 if total accident count $>$ city median, else 0.  
    Tests the ability to separate high‐accident vs.\ low‐accident districts.

  \item[\texttt{accident\_volume\_level}]  
    \hfill \\
    \emph{Three‐class classification:} split total accidents into Low/Medium/High tiers  
    by the 33\% and 67\% quantiles, labeled 0/1/2.  Assesses gradated accident volume encoding.

  \item[\texttt{severity\_entropy}]  
    \hfill \\
    \emph{Binary classification:} compute Shannon entropy of severity‐level proportions  
    \(\{p_i\}_{i=1}^4\), then label = 1 if entropy is greater than median, else 0.  Measures embedding of severity diversity.

  \item[\texttt{has\_sev3}] and \texttt{has\_sev4}  
    \hfill \\
    \emph{Binary classification:}  
    \begin{itemize}
      \item \texttt{has\_sev3}: label = 1 if at least one Severity‐3 accident occurred, else 0.  
      \item \texttt{has\_sev4}: label = 1 if at least one Severity‐4 accident occurred, else 0.  
    \end{itemize}
    Evaluates detection of any serious crashes independently of total counts.

  \item[\texttt{risk\_level}]  
    \hfill \\
    \emph{Three‐class ordinal classification:} combine accident volume and severe‐accident ratio:
    \[
      \text{label} = 
      \begin{cases}
        2 & \text{if volume $>$ median \emph{and} (\(sev3+sev4\))/total $>$ median},\\
        0 & \text{if volume $\leq$ median \emph{and} (\(sev3+sev4\))/total $\leq$ median},\\
        1 & \text{otherwise}.
      \end{cases}
    \]
    Captures joint severity–volume risk levels.
\end{description}

For each task, we extract graph‐level embeddings from each encoder and perform repeated stratified 10‐fold cross‐validation using a linear SVM.  Reported metrics are mean accuracy $\pm$ standard deviation over 1000 repeats.  

\subsection{City-averaged traffic-safety results}

To assess the real-world applicability of FractalGCL, we followed the network collection procedure in \citep{zhai2025heterogeneous} to construct urban road graphs for Chicago, New York, and San Francisco. We then randomly sampled square subgraphs from each complete road network and evaluated six traffic-safety classification tasks following \citep{zhao2024exploring}.

\begin{table}[ht]
\vspace{-0.2cm}
\centering
\caption{Classification accuracy on traffic-safety tasks averaged across Chicago, San Francisco, and New York.}
\label{tab:city}
\resizebox{\linewidth}{!}{
\begin{tabular}{lcccccc}
\toprule
Task & DGI & InfoGraph & GCL & JOAO & SimGRACE & \textbf{FractalGCL} \\
\midrule
total\_accidents\_high & 62.69$\pm$10.80 & 62.33$\pm$10.64 & 64.38$\pm$12.65 & 63.22$\pm$12.41 & 63.56$\pm$12.80 & \textbf{70.30$\pm$13.58} \\
accident\_volume\_level & 48.15$\pm$11.67 & 48.91$\pm$11.71 & 50.52$\pm$12.76 & 48.26$\pm$12.11 & 50.09$\pm$12.73 & \textbf{56.35$\pm$13.68} \\
severity\_entropy & 53.27$\pm$9.24 & 53.17$\pm$9.92 & 53.84$\pm$12.82 & 52.93$\pm$12.02 & 53.16$\pm$12.75 & \textbf{57.44$\pm$14.22} \\
has\_sev3 & 76.67$\pm$12.73 & 75.59$\pm$11.58 & 76.59$\pm$10.66 & 76.26$\pm$11.38 & 76.63$\pm$11.11 & \textbf{80.94$\pm$10.39} \\
has\_sev4 & 54.99$\pm$11.69 & 55.42$\pm$12.08 & 55.75$\pm$13.29 & 54.94$\pm$13.16 & 55.20$\pm$13.38 & \textbf{56.73$\pm$14.08} \\
risk\_level & 40.96$\pm$11.41 & 41.02$\pm$11.29 & 42.44$\pm$13.26 & 40.65$\pm$12.40 & 41.21$\pm$12.91 & \textbf{48.81$\pm$14.43} \\
\midrule
Average & 56.12$\pm$11.26 & 56.07$\pm$11.20 & 57.25$\pm$12.57 & 56.04$\pm$12.25 & 56.64$\pm$12.61 & \textbf{61.76$\pm$13.40} \\
\bottomrule
\end{tabular}}
\end{table}

Table~\ref{tab:city} summarizes six downstream traffic-safety tasks after averaging the city-level results over Chicago, San Francisco, and New York.
FractalGCL attains the highest averaged accuracy on all six tasks and lifts the overall average to \textbf{61.76\%}, a \textbf{4.51} percentage-point improvement over the next-best averaged baseline, GCL (57.25\%).
Full city-level results are reported in Table~\ref{tab:traffic_tasks_pct}.
\begin{table*}[ht]
\centering
\caption{Performance (mean $\pm$ std) on traffic-safety tasks; numbers are in percentage points.}
\label{tab:traffic_tasks_pct}
\scriptsize
\resizebox{\linewidth}{!}{%
\begin{tabular}{llcccccc}
\toprule
Task & City & DGI & InfoGraph & GCL & JOAO & SimGRACE & FractalGCL \\
\midrule
\multirow{3}{*}{total\_accidents\_high} & Chicago & 57.92$\pm$11.57 & 59.26$\pm$12.84 & 63.24$\pm$14.41 & 61.11$\pm$13.32 & 61.82$\pm$14.48 & \textbf{65.48$\pm$14.66} \\
 & SF & 77.28$\pm$13.23 & 78.23$\pm$12.18 & 78.69$\pm$11.93 & 78.60$\pm$13.00 & 79.01$\pm$12.03 & \textbf{79.02$\pm$12.00} \\
 & NY & 52.88$\pm$7.61 & 49.49$\pm$6.89 & 51.20$\pm$11.61 & 49.94$\pm$10.91 & 49.84$\pm$11.89 & \textbf{66.40$\pm$14.07} \\
\midrule
\multirow{3}{*}{accident\_volume\_level} & Chicago & 42.67$\pm$12.00 & 42.73$\pm$12.36 & 45.23$\pm$13.12 & 42.38$\pm$12.17 & 44.42$\pm$13.14 & \textbf{50.41$\pm$13.64} \\
 & SF & 63.82$\pm$12.80 & 66.77$\pm$12.88 & 68.08$\pm$12.56 & 66.62$\pm$12.40 & 68.74$\pm$12.54 & \textbf{68.92$\pm$12.90} \\
 & NY & 37.96$\pm$10.21 & 37.24$\pm$9.89 & 38.25$\pm$12.59 & 35.79$\pm$11.77 & 37.10$\pm$12.51 & \textbf{49.72$\pm$14.51} \\
\midrule
\multirow{3}{*}{severity\_entropy} & Chicago & 51.90$\pm$9.45 & 51.36$\pm$9.68 & 53.28$\pm$12.95 & 51.19$\pm$10.55 & 50.87$\pm$11.74 & \textbf{59.11$\pm$14.84} \\
 & SF & 51.94$\pm$7.56 & 52.69$\pm$9.46 & \textbf{53.37$\pm$11.92} & 52.17$\pm$11.75 & 53.23$\pm$12.02 & 52.54$\pm$13.40 \\
 & NY & 55.96$\pm$10.72 & 55.45$\pm$10.63 & 54.86$\pm$13.60 & 55.43$\pm$13.77 & 55.39$\pm$14.48 & \textbf{60.67$\pm$14.43} \\
\midrule
\multirow{3}{*}{has\_sev3} & Chicago & 77.60$\pm$16.10 & 77.83$\pm$15.42 & \textbf{79.19$\pm$12.56} & 78.44$\pm$13.75 & 78.63$\pm$13.26 & 78.80$\pm$12.35 \\
 & SF & 92.35$\pm$9.75 & 91.41$\pm$5.98 & 93.19$\pm$5.41 & \textbf{93.27$\pm$5.76} & 93.01$\pm$5.42 & 93.13$\pm$5.44 \\
 & NY & 60.05$\pm$12.34 & 57.54$\pm$13.35 & 57.38$\pm$14.02 & 57.07$\pm$14.64 & 58.26$\pm$14.66 & \textbf{70.89$\pm$13.38} \\
\midrule
\multirow{3}{*}{has\_sev4} & Chicago & 53.68$\pm$11.60 & 53.17$\pm$11.74 & 52.91$\pm$13.58 & 53.14$\pm$12.52 & 51.75$\pm$13.15 & \textbf{56.54$\pm$15.00} \\
 & SF & 59.52$\pm$12.93 & 61.14$\pm$13.34 & \textbf{61.84$\pm$13.37} & 59.46$\pm$13.97 & 61.83$\pm$13.69 & 61.53$\pm$14.27 \\
 & NY & 51.77$\pm$10.54 & 51.96$\pm$11.17 & \textbf{52.49$\pm$12.92} & 52.21$\pm$13.00 & 52.01$\pm$13.29 & 52.13$\pm$12.98 \\
\midrule
\multirow{3}{*}{risk\_level} & Chicago & 35.43$\pm$8.52 & 35.75$\pm$8.97 & 38.05$\pm$12.15 & 35.35$\pm$9.86 & 35.78$\pm$11.39 & \textbf{46.22$\pm$14.29} \\
 & SF & 46.23$\pm$13.58 & 48.04$\pm$13.54 & 48.28$\pm$14.24 & 47.60$\pm$14.45 & 47.98$\pm$14.14 & \textbf{48.60$\pm$14.26} \\
 & NY & 41.22$\pm$12.14 & 39.26$\pm$11.35 & 40.98$\pm$13.40 & 39.01$\pm$12.89 & 39.86$\pm$13.21 & \textbf{51.60$\pm$14.74} \\
\midrule
\textbf{Average (\%)} & -- & 56.12$\pm$11.26 & 56.07$\pm$11.20 & 57.25$\pm$12.57 & 56.04$\pm$12.25 & 56.64$\pm$12.61 & \textbf{61.76$\pm$13.40} \\
\bottomrule
\end{tabular}}
\end{table*}

\subsection{Conclusion}

Table~\ref{tab:traffic_tasks_pct} presents the classification accuracies (mean $\pm$ std) of six traffic-safety tasks (total-accidents-high, accident-volume-level, severity-entropy, has\_sev3, has\_sev4, risk\_level) across Chicago, San Francisco, and New York.
The table reports the six methods available in this experiment: DGI, InfoGraph, GCL, JOAO, SimGRACE, and FractalGCL.
FractalGCL achieves the best overall average accuracy of \textbf{61.76\%} and obtains the best score in 13 of the 18 city-task settings.

FractalGCL consistently improves over established contrastive baselines on average across the traffic-safety benchmarks.

We anticipate that FractalGCL’s flexible embedding framework will extend effectively to more complex spatiotemporal and multi-modal urban analytics tasks, such as dynamic traffic flow prediction and integrated land-use and mobility modeling.

\section{Additional Discussion and Limitations}
\label{appendix:discussion}

\subsection{Fractal assumption, applicability, and safe fallback}

FractalGCL is built around a specific inductive bias, namely approximate multi-scale
self-similarity of graphs.
We do \textbf{not} assume that all graphs are fractal; instead, FractalGCL is intended for domains
where the log--log box-counting regression exhibits a clear scaling regime.
In the current version this assumption was mostly implicit in the technical sections, and we make it
explicit here.

Section~3.4 (``Safe fallback under weak fractality or small diameters'') introduces a conservative
gate.
For each graph $G$ we perform the log--log regression and obtain an $R^2(G)$ value, even when the
diameter is small.
In the actual FractalGCL training pipeline, the fractal machinery (renormalised view and fractal
weighting) is activated only when $\mathrm{diam}(G)$ is sufficiently large and $R^2(G)$ exceeds a
threshold; otherwise the gate
\[
  \text{if } \mathrm{diam}(G) \le 9 \text{ or } R^2(G) < \theta, \text{ then disable the fractal loss}
\]
is applied.
In this case the loss reduces exactly to the underlying GCL objective with the same encoder and
augmentations as the baseline.
Thus, on graphs that do not exhibit clear fractal scaling, FractalGCL is designed to behave like a
standard GCL model rather than to harm performance.

Honestly, we do acknowledge that the TU datasets are not collections of very large graphs, so they are not an ideal benchmark for testing FractalGCL; this is precisely why we also included the urban traffic datasets.
Nevertheless, even on these non-large-graph benchmarks, FractalGCL achieves strong performance, which we find encouraging.

\subsection{Finite-graph stability of the estimator and hyperparameters}

The low-complexity estimator used in FractalGCL is designed to operate in the regime where the
fractal assumption holds and the graph diameter is sufficiently large.
We do not claim uniform accuracy across all possible graph scales and topologies, but several parts
of the paper are devoted to controlling the finite-graph behaviour of the estimator.

Theoretically, Theorem~\ref{appendix: estimator_consistency} justifies the finite-scale box-counting estimator under a scaling-limit assumption, and Theorem~\ref{appendix: theorem_renormalisation_keep_dimension} shows that bounded renormalisation preserves the limiting box dimension.
Lemma~\ref{appendix: lemma:_diameter_infinity}, Lemma~\ref{appendix: lemma_ols_clt}, and Proposition~\ref{appendix: proposition_gaussian_surrogate} then motivate the diameter-controlled Gaussian surrogate for the finite renormalisation discrepancy.
This justifies the use of \(\widetilde{\Delta}(G)\) in place of repeatedly recomputing \(\widetilde{\dim}_{\mathrm B}(\mathscr R(G))\) during training.

Algorithmically, we deliberately mitigate the effect of estimator noise in several ways.
First, as discussed above, we apply a conservative gate based on the $R^2$ statistic and the
diameter, so that graphs with unreliable regression never activate the fractal loss and revert to
the baseline GCL objective.
Second, the fractal weight depends on the \emph{difference} of two estimated dimensions and enters
the loss through a smooth exponential factor, which makes the loss relatively insensitive to small
perturbations in $\widehat{\dim}_B$.
Third, Table~3 includes an ablation comparing a variant that uses exact box dimensions
(``w/ Exact Dimension'') with our Gaussian surrogate; the downstream performance is essentially
unchanged while the surrogate reduces computation by about $61\%$, which provides empirical
evidence that FractalGCL is robust to the estimation noise under our current settings.

The choice of renormalisation radius in Algorithm~2 is another potential source of sensitivity.
In practice, we fix $r=1$ in all experiments and include a ``+ random radius'' variant in the
ablation to probe its effect.
Theoretically, in the infinite-graph limit any finite radius is admissible for renormalisation, but
on finite graphs we prefer smaller $r$: a small radius preserves more local information while still
enabling meaningful coarse-graining.
Empirically, we observe that using random radii (which occasionally produce much larger effective
$r$) tends to degrade accuracy, consistent with the intuition that overly large boxes make the
renormalised graph too coarse.
Taken together, these observations suggest that $r=1$ is a conservative and relatively robust
choice in our current setting.

\subsection{Experimental scope and computational trade-offs}

The present work focuses on graph-level representation learning, using TU molecule/protein
benchmarks and urban traffic networks.
This choice is intentional: our theoretical development (renormalisation, box dimension, dimension
gap) is formulated at the graph level, and all experiments are designed to match this setting.
We acknowledge that our current work does not include node-level or link-level experiments or
theory; intuitively, we expect that global graph-level signals such as fractal scaling and
multi-scale self-similarity could also benefit node-level objectives (for example by guiding message
passing or regularising node embeddings), and we see extending FractalGCL to node-level tasks as a
promising direction for future work.

For contrastive learning, our setup uses the standard in-batch negative protocol used by many graph-level GCL methods: the positive pair is formed from the anchor graph and its augmented view, while the negative set consists of augmented views from other graphs in the same minibatch.
FractalGCL does not change which graphs are sampled as negatives; instead, it changes how strongly those fixed negatives contribute to the denominator through a finite-scale box-counting discrepancy.
This should be understood as a multi-scale structural prior rather than a claim that labels or semantic similarity are determined by box-counting dimension alone.
The strength of the prior is controlled by \(\alpha\), and the fractal branch is disabled when the box-counting fit is weak or the graph diameter is too small.
A systematic comparison with methods that explicitly mine hard negatives or correct contrastive sampling bias would be valuable, and we view such approaches as complementary to our contribution.

Regarding computational cost, FractalGCL is indeed more involved than the simplest GCL baselines,
mainly because it requires computing graph diameters and box dimensions.
We view this as a natural trade-off between computational cost and representational power: some
additional preprocessing cost is inevitable if one wishes to enforce global structural constraints
and go beyond purely heuristic augmentations.
In practice, diameters and box dimensions are computed once per graph as offline preprocessing and
cached; during training the Gaussian surrogate adds only an $O(N^2)$ sampling cost on top of the
similarity matrix already present in InfoNCE.
The ablation study shows that replacing exact box-counting by the surrogate reduces the exact-dimension runtime from 1249.74 s to 486.81 s, corresponding to a 2.56-fold speed-up or about a 61\% runtime reduction, while maintaining essentially the same accuracy.

\section{Conclusion}\label{sec:conclusion}

We presented FractalGCL, a graph contrastive learning framework that incorporates fractal geometry through renormalisation-based augmented views and a fractal-dimension-aware contrastive objective.
The central idea is to move beyond purely local perturbations by using finite-scale box-counting information to guide both view construction and contrastive weighting.
Empirically, FractalGCL serves as an effective frozen-pretraining representation on MalNet-Tiny, achieves strong graph-level representation performance on the TUDataset benchmarks considered in this work, and obtains the highest average accuracy among the compared baselines on real-world urban traffic tasks.
At the same time, the Gaussian surrogate makes the dimension-correction step substantially more efficient, reducing the runtime from 1249.74 s under exact computation to 486.81 s, which corresponds to a 2.56-fold speed-up and about a 61\% runtime reduction.

A limitation of FractalGCL is that its benefits depend on the presence of meaningful finite-scale fractal structure.
When a graph has weak fractal scaling or very small diameter, the fractal correction may become less informative, and the method should rely more heavily on its fallback mechanism.
Future work will benchmark FractalGCL across datasets with a wider range of fractal degrees, develop sharper finite-graph diagnostics for the renormalisation step, and study when fractal-dimension-aware objectives are most useful.
Overall, our results suggest that fractal geometry provides a useful perspective for designing graph representation learning methods on networks with non-trivial multiscale structure.

\newpage
\section*{NeurIPS Paper Checklist}

\begin{enumerate}

\item {\bf Claims}
    \item[] Question: Do the main claims made in the abstract and introduction accurately reflect the paper's contributions and scope?
    \item[] Answer: \answerYes{}
    \item[] Justification: The abstract and introduction state the scope of the method: a renormalisation-based augmented graph, a finite-scale box-counting loss, a Gaussian surrogate for renormalisation discrepancy, and a reliability gate. The theoretical results are in Section~\ref{subsection:background-of-fractal-learning}--Section~\ref{subsection:implementation}, and the empirical claims are supported in Section~4 and the appendices.

\item {\bf Limitations}
    \item[] Question: Does the paper discuss the limitations of the work performed by the authors?
    \item[] Answer: \answerYes{}
    \item[] Justification: The paper discusses applicability and limitations through the reliability gate in Section~\ref{subsection:safe-fallback} and through the additional discussion on fractal assumptions, finite-graph stability, hyperparameters, experimental scope, and computational trade-offs in Appendix~\ref{appendix:discussion}.

\item {\bf Theory assumptions and proofs}
    \item[] Question: For each theoretical result, does the paper provide the full set of assumptions and a complete (and correct) proof?
    \item[] Answer: \answerYes{}
    \item[] Justification: The theoretical assumptions and statements are given in Section~\ref{subsection:background-of-fractal-learning}--Section~\ref{subsection:implementation}. Complete proofs for the box-counting estimator, renormalisation theorem, loss-gradient result, complexity bound, and Gaussian surrogate are provided in Appendix~\ref{appendix: math}.

    \item {\bf Experimental result reproducibility}
    \item[] Question: Does the paper fully disclose all the information needed to reproduce the main experimental results of the paper to the extent that it affects the main claims and/or conclusions of the paper (regardless of whether the code and data are provided or not)?
    \item[] Answer: \answerYes{}
    \item[] Justification: The experimental sections describe the datasets, frozen-pretraining protocol, TUDataset evaluation protocol, urban-traffic evaluation, hyperparameters, and hardware. The released anonymized code repository is intended to provide the scripts needed to reproduce the reported experiments.

\item {\bf Open access to data and code}
    \item[] Question: Does the paper provide open access to the data and code, with sufficient instructions to faithfully reproduce the main experimental results, as described in supplemental material?
    \item[] Answer: \answerYes{}
    \item[] Justification: The paper provides an anonymized code link and uses public benchmark datasets such as TUDataset and MalNet-Tiny. Details for data processing and experiment construction are described in the experimental sections and appendices.

\item {\bf Experimental setting/details}
    \item[] Question: Does the paper specify all the training and test details (e.g., data splits, hyperparameters, how they were chosen, type of optimizer) necessary to understand the results?
    \item[] Answer: \answerYes{}
    \item[] Justification: The paper reports the key experimental settings, including architectures, pretraining and evaluation protocols, cross-validation, hyperparameters, optimizer choices, and dataset splits in Section~4 and the appendices.

\item {\bf Experiment statistical significance}
    \item[] Question: Does the paper report error bars suitably and correctly defined or other appropriate information about the statistical significance of the experiments?
    \item[] Answer: \answerYes{}
    \item[] Justification: The main experimental tables report means and standard deviations where repeated runs, folds, or seeds are used. The paper specifies repeated 10-fold cross-validation for TUDataset and three-seed averaging for the MalNet-Tiny frozen-pretraining experiment.

\item {\bf Experiments compute resources}
    \item[] Question: For each experiment, does the paper provide sufficient information on the computer resources (type of compute workers, memory, time of execution) needed to reproduce the experiments?
    \item[] Answer: \answerYes{}
    \item[] Justification: The paper reports the compute environment for the main experiments, including the use of an NVIDIA A100 GPU and runtime information for the MalNet-Tiny experiment. Additional implementation and hyperparameter details are provided in the experimental sections.
    
\item {\bf Code of ethics}
    \item[] Question: Does the research conducted in the paper conform, in every respect, with the NeurIPS Code of Ethics \url{https://neurips.cc/public/EthicsGuidelines}?
    \item[] Answer: \answerYes{}
    \item[] Justification: The research uses public graph benchmarks and methodological experiments, and we are not aware of any deviation from the NeurIPS Code of Ethics.

\item {\bf Broader impacts}
    \item[] Question: Does the paper discuss both potential positive societal impacts and negative societal impacts of the work performed?
    \item[] Answer: \answerYes{}
    \item[] Justification: The paper discusses potential positive impacts in graph pretraining, malware analysis, and traffic-safety modelling, while also noting possible risks related to misuse, privacy, fairness, and domain-specific deployment.
    
\item {\bf Safeguards}
    \item[] Question: Does the paper describe safeguards that have been put in place for responsible release of data or models that have a high risk for misuse (e.g., pre-trained language models, image generators, or scraped datasets)?
    \item[] Answer: \answerNA{}
    \item[] Justification: The paper does not release a high-risk model or dataset such as a pretrained language model, image generator, or scraped unsafe-content dataset. The released asset is an anonymized implementation of a graph representation learning method.

\item {\bf Licenses for existing assets}
    \item[] Question: Are the creators or original owners of assets (e.g., code, data, models), used in the paper, properly credited and are the license and terms of use explicitly mentioned and properly respected?
    \item[] Answer: \answerYes{}
    \item[] Justification: The paper credits the public datasets, benchmark methods, and open-source software packages used in the experiments, and documents their corresponding licences and terms of use in the released anonymized repository.

\item {\bf New assets}
    \item[] Question: Are new assets introduced in the paper well documented and is the documentation provided alongside the assets?
    \item[] Answer: \answerYes{}
    \item[] Justification: The paper introduces and releases a new method implementation through an anonymized repository. The repository and paper describe the main algorithmic components, hyperparameters, and experiment protocols.

\item {\bf Crowdsourcing and research with human subjects}
    \item[] Question: For crowdsourcing experiments and research with human subjects, does the paper include the full text of instructions given to participants and screenshots, if applicable, as well as details about compensation (if any)? 
    \item[] Answer: \answerNA{}
    \item[] Justification: The work does not involve crowdsourcing experiments or research with human subjects.

\item {\bf Institutional review board (IRB) approvals or equivalent for research with human subjects}
    \item[] Question: Does the paper describe potential risks incurred by study participants, whether such risks were disclosed to the subjects, and whether Institutional Review Board (IRB) approvals (or an equivalent approval/review based on the requirements of your country or institution) were obtained?
    \item[] Answer: \answerNA{}
    \item[] Justification: The work does not involve crowdsourcing experiments or research with human subjects, so IRB approval or an equivalent review is not applicable.

\item {\bf Declaration of LLM usage}
    \item[] Question: Does the paper describe the usage of LLMs if it is an important, original, or non-standard component of the core methods in this research? Note that if the LLM is used only for writing, editing, or formatting purposes and does \emph{not} impact the core methodology, scientific rigor, or originality of the research, declaration is not required.
    \item[] Answer: \answerNA{}
    \item[] Justification: LLMs are not used as part of the core method, experiments, or scientific contribution. Any language editing or formatting assistance is outside the core methodology and does not affect the originality of the research.

\end{enumerate}

\end{document}